\documentclass[10pt,twocolumn,letterpaper]{article}

\usepackage{iccv}
\usepackage{times}
\usepackage{epsfig}
\usepackage{graphicx}
\usepackage{amsmath}
\usepackage{amssymb} 

\newcommand{\COMMENT}[1]{}

\usepackage[pagebackref=true,breaklinks=true,letterpaper=true,colorlinks,bookmarks=false]{hyperref}
\usepackage[all]{hypcap}

\iccvfinalcopy 

\def\iccvPaperID{1128} 

\newenvironment{tightitemize} 
{\vspace{-\topsep}\begin{itemize}\itemsep1pt \parskip0pt \parsep0pt}
	{\end{itemize}\vspace{-\topsep}}

\usepackage{multirow}

\usepackage{calc,advdate,datetime}

\usepackage[outline]{contour}
\usepackage[export]{adjustbox}

\begin{document}


\title{Beyond Planar Symmetry: \\
Modeling human perception of reflection and rotation symmetries in the wild}
\author{Christopher Funk \qquad\qquad Yanxi Liu\\
	School of Electrical Engineering and Computer Science. \\ 
	The Pennsylvania State University.  
	University Park, PA.  16802, USA \\
	{\tt\small \{\href{mailto:funk@cse.psu.edu}{funk}, \href{mailto:yanxi@cse.psu.edu}{yanxi}\}@cse.psu.edu}}
\maketitle
%
\begin{abstract}
Humans take advantage of real world symmetries for various tasks, yet capturing their superb symmetry perception mechanism 
with a computational model remains elusive. 
Motivated by a new study demonstrating the extremely high inter-person accuracy of human perceived symmetries in the wild, we have constructed the first deep-learning neural network for reflection and rotation symmetry detection (Sym-NET), 
trained on photos from MS-COCO (Microsoft-Common Object in COntext) dataset with nearly 11K consistent symmetry-labels from more than 400 human observers. 
We employ novel methods to convert discrete human labels into symmetry heatmaps, capture symmetry densely in an image and quantitatively evaluate Sym-NET against multiple existing computer vision algorithms. 
On CVPR 2013 symmetry competition testsets and unseen MS-COCO photos, Sym-NET significantly outperforms all other competitors. Beyond mathematically well-defined symmetries on a plane, Sym-NET demonstrates abilities to identify viewpoint-varied 3D symmetries, partially occluded symmetrical objects, and symmetries at a semantic level.

\end{abstract}

\section{Introduction}
%
From the evolution of plants, insects and mammals, as well as the earliest pieces of art in 20,000 BCE through the modern day \cite{Giurfa1996,jones1910,thompson1961},
perfectly symmetrical objects and scenes are rare while approximate symmetries are readily observable in both natural and man-made worlds. 
Perception of such {\em symmetries in the wild} has played an instrumental role 
at different levels of intelligence 
\cite{delius1978symmetry,driver1992preserved,Giurfa1996,hafting2005microstructure,parovel2002mirror,rodrlguez2004,scheib1999facial,von1992dolphin} that function effectively in an otherwise cluttered and often uncertain world. 
For decades, among human vision and computer vision researchers alike, the search for 
computational models, analytical or biological basis, and explanations
for symmetry perception  \cite{kohler2016representation,leyton1992,treder2010behind,tyler1996} has proven to be non-trivial \cite{biederman1985cvgip,liu_etal2010,tyler2002}. 

\COMMENT{
From the evolution of plants, insects and mammals, and from the earliest pieces of art in 20,000 BCE through the modern day \cite{Giurfa1996,jones1910,thompson1961},
perfectly symmetrical objects and scenes are rare, while approximate symmetries are readily observable in both natural and man-made worlds. 
Perception of such {\em symmetries in the wild} has played an instrumental role 
at different levels of intelligence, from insects to humans 
\cite{biederman1985cvgip,delius1978symmetry,Giurfa1996,hafting2005microstructure,rodrlguez2004,von1992dolphin}, 
that function effectively in an otherwise cluttered, dynamic and often uncertain world. 
Among human vision and computer vision researchers alike, the search for 
computational basis and explanations for symmetry 
perception  \cite{kohler2016representation,leyton1992,treder2010behind,tyler1996} has proven to be non-trivial \cite{liu_etal2010,tyler2002}. 
}
\begin{figure}[!t] 
	\centering
	\newlength{\figOneWidth}
	\setlength{\figOneWidth}{.295\linewidth}
	\setlength{\tabcolsep}{4pt} 
	\begin{tabular}{ccc}
			Original Image & Symmetry GTs & Predicted \\
		\includegraphics[width=\figOneWidth]{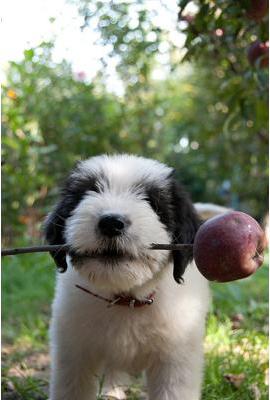} &
		\includegraphics[width=\figOneWidth]{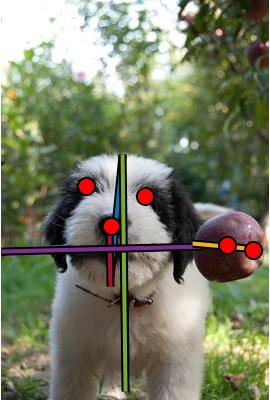}  
		&
		\includegraphics[width=\figOneWidth]{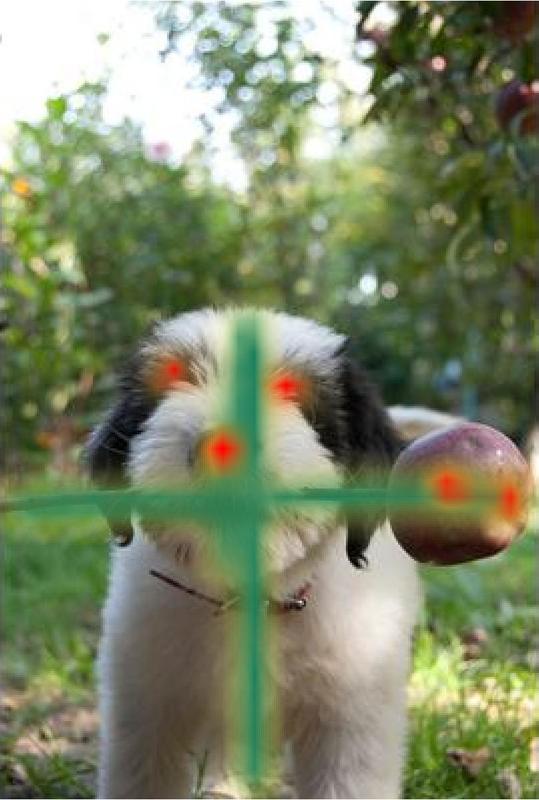} \\
		\includegraphics[width=\figOneWidth]{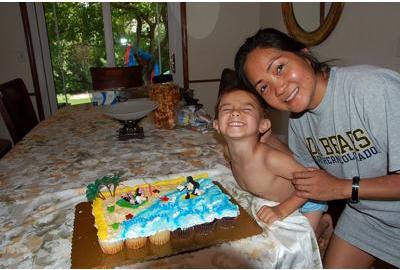} &
		\includegraphics[width=\figOneWidth]{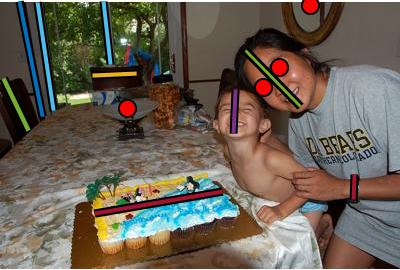}  
		&
		\includegraphics[width=\figOneWidth]{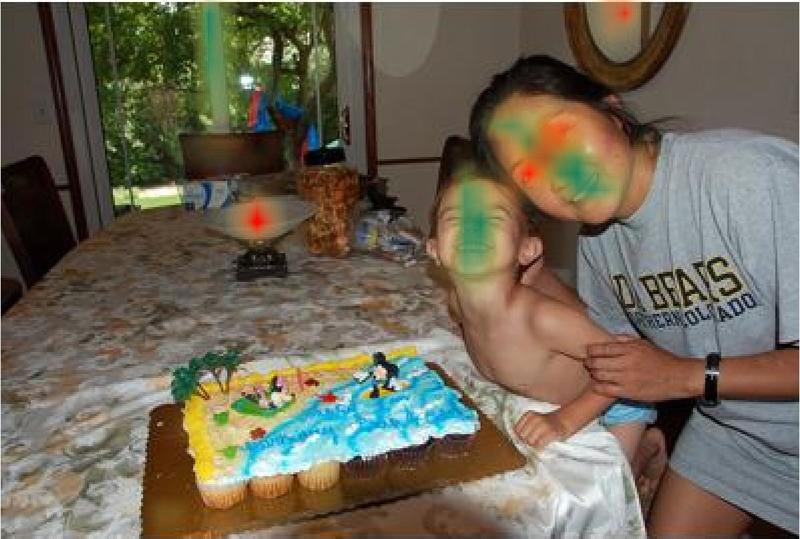} \\
		\includegraphics[width=\figOneWidth]{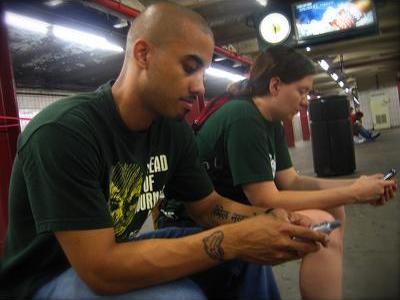} &
		\includegraphics[width=\figOneWidth]{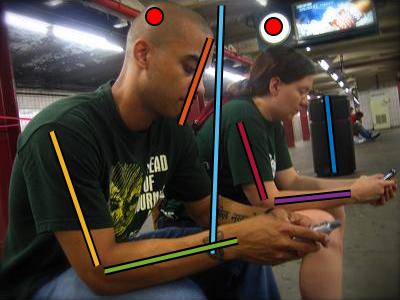} 
		&
		\includegraphics[width=\figOneWidth]{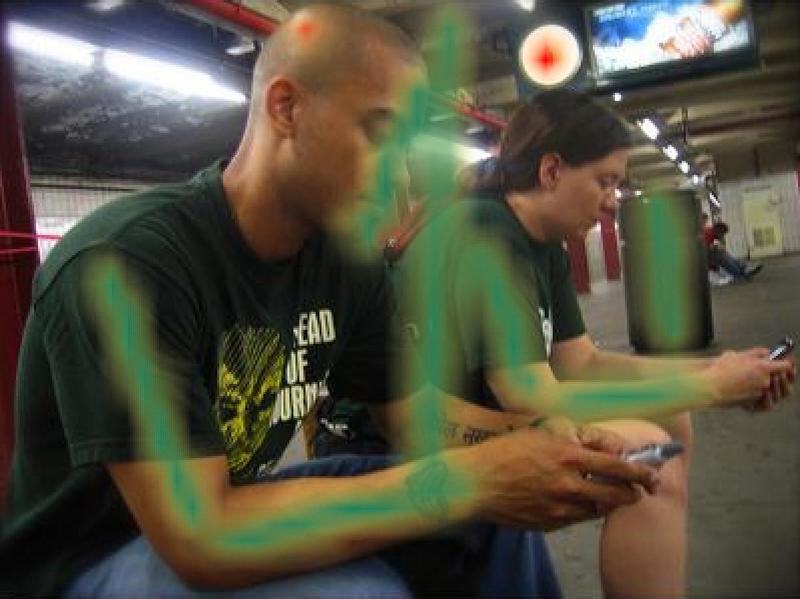}
	\end{tabular}
	\setlength{\tabcolsep}{6pt} 
	\caption{
	Sample 
	images from the MS-COCO dataset~\cite{lin2014microsoft}.  
	Symmetry ground-truths (mid-column) are computed from human labels (Figure \ref{fig:Max Number of Labeled Symmetries}):
	line segments for reflection symmetry axes and red dots for rotation symmetry centers.
	Right column: heatmaps for predicted symmetries, \textcolor{green}{green} for reflection symmetry axes, and 
\textcolor{red}{red} for rotation symmetry centers.
	}
	\label{fig:symmetry basics}
	\vspace{-15pt}
\end{figure}

The mathematical definition of a symmetry transformation $g$ of a set $S$ is clear and simple \cite{coxeter1980,liu_etal2010,weyl1952}, $g(S) = S$. 
However, how to identify a symmetry in a photo remains ambiguous
 \cite{liu2013symmetry,rauschert2011symmetry}.
The dilemma is: should a symmetry $g$ in an image be determined by its 
mathematical definition or human perception? 
Human perception of symmetry can deviate grossly from 2D pixel-geometry on an image.
In Figure~\ref{fig:symmetry basics}: a face profile is perceived as having a reflection symmetry; 
a rotation center is identified for half of a round mirror, and for the highly skewed top-center of an apple;
and a reflection symmetry is labeled between two sitting people playing with their phones!
None of this mixture of 3D/object prior-based and semantic-level symmetries has been attempted in existing symmetry detection models.

Motivated by the strong human perceptual consistency of symmetry demonstrated in a recent study on 
1,200 MS-COCO \cite{lin2014microsoft} images rated by 400 human raters~\cite{funk_liuCVPR2016Symmetry}, we take a first step in building a computational platform for learning to mimic human visual perception of reflection and rotation symmetries.
Though multi-layer Convolutional Neural Networks (CNNs) have been trained to detect image class~\cite{krizhevsky2012imagenet,simonyan2014very,szegedy2015going}, semantic segments~\cite{chen2014semantic,chen2016deeplab,girshick2014rich,long2015fully}, surface normals~\cite{wang2015designing}, face identities~\cite{taigman2014deepface}, human pose~\cite{cao2016realtime,pfister2015flowing,wei2016convolutional},  
and to be invariant to rotational symmetry~\cite{dieleman2016exploiting,laptev2016ti,worrall2016harmonic}, little has been reported on training CNNs for reflection and rotation symmetry detection on real images.  %
%
%
Using state-of-the-art segmentation networks as a base~\cite{chen2016deeplab},
we transform ground truth extracted from human labels 
into dense 2D symmetry heatmaps (as opposed to sparse labels containing only 2D coordinates), and perform dense (per pixel) regression to those heatmaps.
We compare against existing algorithms that output symmetry heatmaps with the same dimensions as the input image.  

Our contributions are:
\begin{tightitemize}
	\item to build the first deep, dense, and multiple symmetry detector that mimics sophisticated human symmetry perception beyond planar symmetries;
	\item to convert sparse symmetry labels into dense heatmaps to facilitate CNN training using human labels;
	\item to systematically and extensively validate and compare the predictive power of the trained CNN against existing algorithms on both mathematically well-defined and human perceived symmetries.
\end{tightitemize}
In short, this work advances state-of-the-art on {\em symmetry detection in the wild} by 
using CNNs to effectively mimic human perception of reflection and rotation symmetries.

\section{Related Work}


One can find a general review of human 
symmetry perception (primarily reflection) in \cite{tyler2002}, and on computational aspects of symmetry detection in \cite{liu_etal2010}.

\subsection{Reflection Symmetry Detection}
Reflection symmetry algorithms fall into two different categories depending on whether they detect sparse symmetries
(straight lines or curves) \cite{hauagge2012image,lee2009curved,lee2012curved,liu2010curved,loy2006detecting,widynski2014local} or a dense heatmap~\cite{fukushima2005use,fukushima2006symmetry,tsogkas2012learning}. The most common sparse approach to detect reflection symmetry is to match up symmetric points or contours in the image to determine midpoint and direction of the symmetry axis~\cite{cai2014adaptive,levitt1984domain,loy2006detecting,mignotte2016symmetry,ogawa1991symmetry,wang2015reflection}.  These approaches often use a Hough transform to accumulate the axes of reflection, derived from the matched feature's midpoints and angles, and vote on the dominant symmetries.  Atadjanov and Lee~\cite{atadjanov2015bilateral} extend the Loy and Eklundh~\cite{loy2006detecting} algorithm by taking the matched keypoints and then comparing the histogram of curvature around the keypoints.  Wang~\etal~\cite{wang2015reflection} uses local affine invariant edge correspondences to make the algorithms more resilient to perspective distortion contours.  The method does not use a Hough space to vote, opting instead to use an affine invariant pairwise (dis)similarity metric to vote for symmetries.  

Pritts~\etal~\cite{pritts2014detection} detect reflection, rotation and translation symmetry using SIFT and MSER features.  The symmetries are found through non-linear optimization and RANSAC.
Tuytelaars~\etal~\cite{tuytelaars_etal2003} detect reflection through a Cascade Hough Transform.   Kiryati and Gofman~\cite{kiryati1998detecting} define a Symmetry ID function implemented through Gaussian windows to find local reflection symmetry.

Lee and Liu~\cite{lee2009curved,lee2012curved} 
have generalized the traditional straight reflection axes detection problem into 
finding {\em curved glide-reflection symmetries}.
 Their approach adds a translational dimension in the Hough transform so the matched features are clustered in a 3D parameter space, and the curved reflection or glide-reflection axis is found by polynomial regression between clustered features.

Tsogkas and Kokkinos~\cite{tsogkas2012learning} use a learning based approach for local reflection symmetry detection.  Features are extracted using rotated integrals of patches in a Gaussian pyramid and converted into histograms.  These features are spectrally clustered and multiple instance learning is used to find symmetries with multiple scales and orientations simultaneously. Teo~\etal~\cite{teo2015detection} detect curved-reflection symmetry using structured random forests (SRF) and segment the region around the curved reflection.  The SRF are trained using multi-scale intensity, LAB, oriented Gabor edges, texture, and spectral features.  Many trees are trained and the output of the leaves for the trees are averaged to obtain the final symmetry axes.  

There have been some shallow-network reflection detection approaches (well before the current deep learning craze).  Zielke~\etal~\cite{zielke1992intensity} use a static feed forward method to enhance the symmetric edges for detection.  The max operation between the different orientations is similar to other voting systems~\cite{liu_etal2010}.
Fukushima~and~Kikuchi~\cite{fukushima2005use,fukushima2006symmetry} present another neural network method for detecting reflection symmetry around the center of an image.  
They use a 4-layer network to find the symmetry axis.   

{\em Skeletonization}, a related problem to reflection detection, has attracted a lot of attention recently~\cite{lee2015framework,shen2016multiple,tsogkas2012learning,widynski2014local}.
Shen~\etal~\cite{shen2016object} use a deep CNN to learn symmetry at multiple scales and fuse the final output together.  The network needs object skeleton ground-truth for the particular scale of the objects.  
The network outputs a skeleton heatmap which is thresholded to produce a binary image denoting the detected skeletons.

 \subsection{Rotation Symmetry Detection}
 Earlier work on rotation symmetry detection includes the use of autocorrelation \cite{krahe1986detection,liu_etal2004PAMI} and image moments \cite{chou1991fold,marola1989using,tsai1991detection}.  
%
 %
 Loy and Eklundh~\cite{loy2006detecting} use a variation on their SIFT feature-based reflection symmetry approach to find rotation symmetry as well.  
 The orientations of matched SIFT feature pairs are used to 
 find a rotation symmetry center.  The detected rotation symmetry centers emerge 
 as maxima in the voting space. This algorithm stands out from all others since the authors have made their code publicly available, and the symmetry competition workshops in CVPR 2011/2013 have used it as the baseline algorithm for both reflection and rotation symmetry detection. Thus far, this algorithm is considered 
 to be the best baseline algorithm for reflection and rotation symmetry detection.

 Lee and Liu~\cite{lee2009symmetry,lee2008rotation,lee2009slewed} have developed an algorithm to detect
 (1) the center of rotation, 
 (2) the number of folds,
 (3) type of symmetry group (dihedral/cyclic/O(2)), and
 (4) the region of support.  
 The first step of their algorithm is rotation symmetry center detection where they
 use a novel {\em frieze expansion} to transform the image at each pixel location into polar coordinates and then search for translation symmetry. 
The second step applies a Discrete Fourier Transform (DFT) on the frieze expansion to determine (2)-(4) listed above.
In our work, for rotation symmetries we only focus on detecting rotation symmetry centers.


\subsection{Dense CNN Regression}
Fully Convolutional Networks~\cite{long2015fully}, with CNN regressing to 2D ground-truth, 
have been utilized for  semantic segmentation~\cite{chen2016deeplab,long2015fully} and pose detection~\cite{cao2016realtime,pfister2015flowing,wei2016convolutional}.  For semantic segmentation, the output of the network is an $ n \times n \times c $ matrix where $ n $ is a reduced dimension of the input image and c is the number of classes.  A pixel-wise argmax operation is computed for each $ n \times n  $ pixel across $ c $ to classify the corresponding class. Chen~\etal~\cite{chen2016deeplab} uses a pyramid of upsampling atrous filters~\cite{chen2014semantic,chen2016deeplab,holschneider1990real,papandreou2015modeling} which enables more context to inform each pixel in the network output.  
A heatmap regression for each joint is estimated separately for human pose detection \cite{cao2016realtime,pfister2015flowing,wei2016convolutional} where a Gaussian is defined over each ground-truth label to provide an easier target to regress.  Without this Gaussian, the only error signaling a correct label would be from the single pixel of ground-truth output and the network would predict everything as background.  The networks are trained with $ \ell_2 $ loss.
We employ the same architecture as Chen~\etal~\cite{chen2016attention} to detect symmetries while using a 2D heatmap regression similar to pose detection.  Both the added context and additionally the multiple scales are relevant in detecting symmetry within the images.  Similar to pose detection, we use an $ \ell_2 $ regression where each ground-truth is defined by a Gaussian centered at the ground-truth label.  

Different from recent efforts in the deep learning/CNN community where researchers are seeking networks that are 
rotation/reflection or affine invariant to input images \cite{dieleman2016exploiting,gens_domingos2014deep,laptev2016ti,worrall2016harmonic}, our work explicitly acknowledges (near) reflection and rotation symmetries in the raw data regardless of the transformations applied on the input images.
To the best of our knowledge, there have been no deep learning networks trained on human symmetry labels 
for automated reflection and rotation symmetry detections.

%
\section{Our Approach}

We propose a multi-layer, fully-convolutional neural network for reflection and rotation symmetry from real world images.  We call this Sym-NET which is short for \textbf{SYM}metry detection neural \textbf{NET}work.  

\subsection{Data and Symmetry Heatmaps}
\begin{table}[b!] \centering \vspace{-10pt}
	\def\columnWidthGT{.64\linewidth}
	\newcommand{\statRow}[2]{
		\parbox[t]{\columnWidthGT}{\centering #1} & #2 \\ \hline
		}
	\begin{tabular}{c|c} \hline
		\statRow{Total \# of Images with GT}{1,202}
		\statRow{Total \# of Images with Reflection GT}{1,199}
		\statRow{Total \# of Images with Rotation GT}{1,057}
		\statRow{Mean \# of GT Labelers  $ \pm $ std/ Image}{$ 29.18 (\pm 4.04) $} 
		\statRow{Mean \# of Reflection GT Labelers$\pm$std / Image}{$ 23.99 (\pm 6.67) $} 
		\statRow{Mean \# of Rotation GT Labelers $\pm$std / Image}{$ 13.00 (\pm 10.33) $} 
		\statRow{Mean \# GT  $ \pm $ std / Image}{$ 9.14 (\pm 4.74  ) $} 
		\statRow{Mean \# Reflection GT $ \pm $ std/Image }{$  6.05 (\pm 3.28) $}
		\statRow{Mean \# Rotation GT $ \pm $ std / Image}{$ 3.09 (\pm 2.93) $} 
		\statRow{Total GT}{10,982} 
		\statRow{Total Reflection GT}{7,273}
		\statRow{Total Rotation GT}{3,709}  
		\statRow{Total \# of Labels}{107,172}
		\statRow{Total \# of Labels Used for GTs}{77,756}  
	\end{tabular} 
	\vspace{1pt} 
	\caption{\label{tab:GT dataset} 
		Statistics of labeled symmetries used in this work for training and testing Sym-NETs.
	} \vspace{-0pt} 
\end{table}

\begin{figure}[!t]
	\centering
		\parbox{.95\linewidth}{\centering \includegraphics[width=\linewidth]{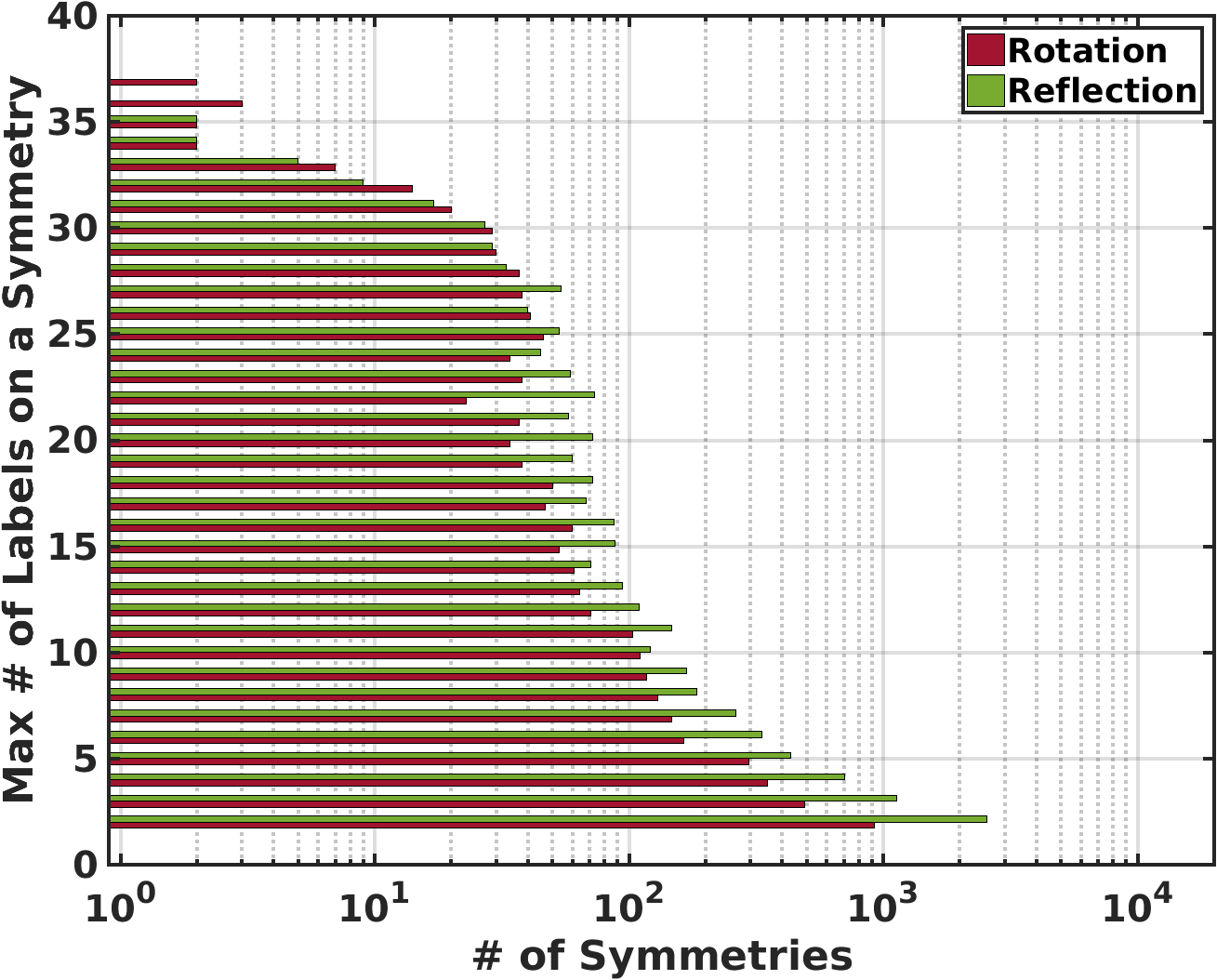}} 
	\caption{Histogram indicating the number of distinct human labelers for \underline{each} labeled reflection and rotation symmetry in the image dataset, respectively.   
	The higher the number on the Y axis, the more consistent the raters
are (agreeing with each other).   }
	\label{fig:Max Number of Labeled Symmetries} \vspace{-10pt}
\end{figure}

The raw data is a collection of images from Microsoft COCO dataset~\cite{lin2014microsoft}.
The symmetry Ground-Truth (GT) labels have been collected from Amazon Mechanical Turk (Table~\ref{tab:GT dataset}).  The data includes reflection axes (two points per axis) and rotation centers (one point per center).  
The statistics of the human labeled symmetries are 
shown in Figure~\ref{fig:Max Number of Labeled Symmetries}.  
We gather human symmetry labels using the same tool as described in 
\cite{funk_liuCVPR2016Symmetry} 
on a superset of images reported there. 
%
%

\label{sec:sparse to dense}

To obtain a
consensus on symmetry GTs computationally, 
we first combine human perceived symmetry labels 
with an automated clustering algorithm~\cite{funk_liuCVPR2016Symmetry}.
The basic idea is to capture the exponential divergence in the nearest labeled symmetry pair distribution, use that as the minimum distance $\tau$ between neighbors and 
the number of required human labels as the minimum number of neighbors, and finally input both to DBSCAN \cite{ester1996density}, a 
method for {\bf D}ensity-{\bf B}ased {\bf S}patial {\bf C}lustering of {\bf A}pplications with {\bf N}oise (the winner of the test-of-time award in 2014). The $\tau$ for rotation symmetry perception is 5 pixels, i.e. 
two symmetry labels within $\tau$ are considered to be labeling the same perceived symmetry \cite{funk_liuCVPR2016Symmetry}.
Second, these sparse symmetry GTs on each image are mapped into a reflection or rotation symmetry heatmap respectively \cite{cao2016realtime,pfister2015flowing,wei2016convolutional}.  
Let $ GT^k $ be all the pixel location(s) ($ l $) for a 1 pixel wide reflection symmetry axis or a 1 pixel rotation symmetry center and let $ x_{i,j} $ be all the pixel locations for the input image.  We create the dense ground truth symmetry heatmap ($H$) for each ground-truth symmetry $ k $ with a $ \sigma $ of 5 (the $\tau$ found in \cite{funk_liuCVPR2016Symmetry}):

\begin{equation}
H_{i,j,k} =\sum_{l \in GT^k} \exp\left(-\frac{||l -x_{i,j}||_2^2}{2\sigma^2}\right).
\end{equation}

\begin{figure}[t!]
	\centering
	\newlength{\figGTWidth}
	\setlength{\figGTWidth}{.1875\linewidth}
	
	\setlength{\tabcolsep}{1pt} 
	\begin{tabular}{ccccc}
		\parbox{\figGTWidth}{\centering Original Image} & 
		\parbox{\figGTWidth}{\centering Human Labels} & 
		\parbox{\figGTWidth}{\centering Symmetry GTs} & 
		\multicolumn{2}{c}{\parbox{2\figGTWidth}{\centering Symmetry Heatmap ($ H $)}} \\
					\COMMENT{
		\includegraphics[width=\figGTWidth]{./results_images/Reflection/im/COCO_test2014_000000004099} &
		\includegraphics[width=\figGTWidth]{./results_images/both/human/COCO_test2014_000000004099} &
		\includegraphics[width=\figGTWidth]{./results_images/both/gt/COCO_test2014_000000004099}  	&
		\includegraphics[width=\figGTWidth]{./results_images/both/gt_heatmap/COCO_test2014_000000004099} 	&
		\includegraphics[width=\figGTWidth]{./results_images/both/gt_heatmap_blank/COCO_test2014_000000004099} 
		\\
		\includegraphics[width=\figGTWidth]{./results_images/Reflection/im/COCO_test2014_000000521507} &
		\includegraphics[width=\figGTWidth]{./results_images/both/human/COCO_test2014_000000521507} &
		\includegraphics[width=\figGTWidth]{./results_images/both/gt/COCO_test2014_000000521507}  &
		\includegraphics[width=\figGTWidth]{./results_images/both/gt_heatmap/COCO_test2014_000000521507}   &
		\\
		}
		\includegraphics[width=\figGTWidth]{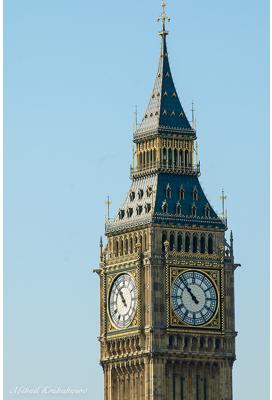} &
		\includegraphics[width=\figGTWidth]{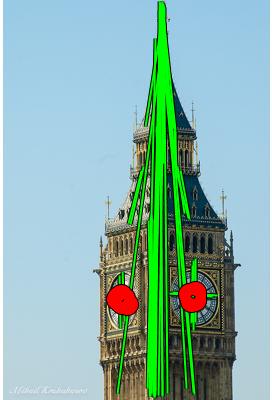} &
		\includegraphics[width=\figGTWidth]{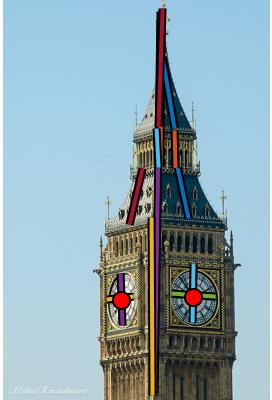}  	&
		\includegraphics[width=\figGTWidth]{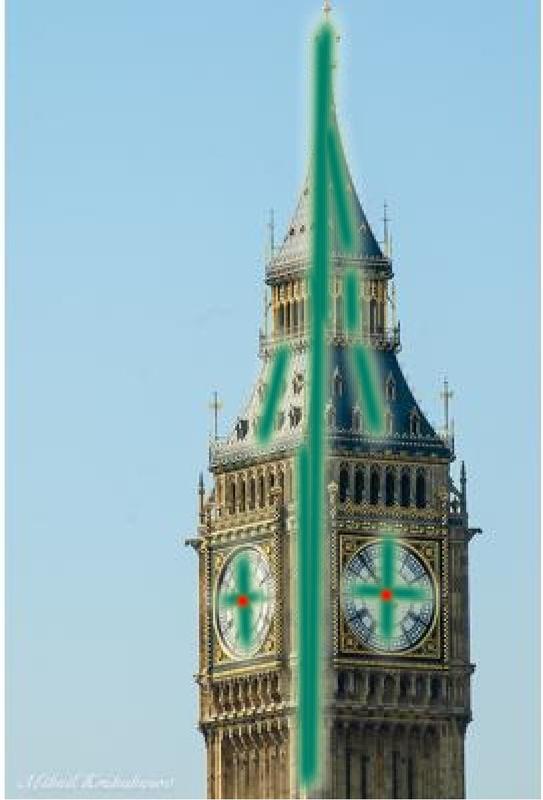} 	&
		\includegraphics[width=\figGTWidth]{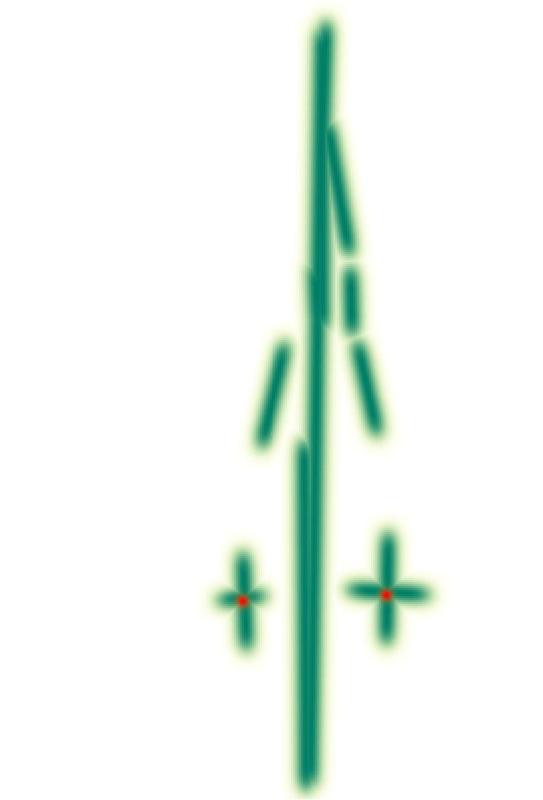} 
		\\
		\includegraphics[width=\figGTWidth]{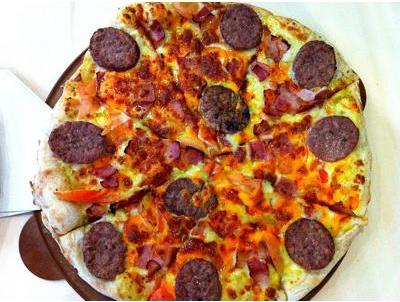} &
		\includegraphics[width=\figGTWidth]{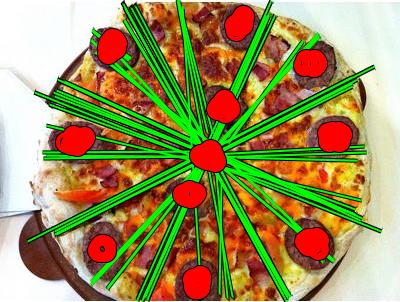} &
		\includegraphics[width=\figGTWidth]{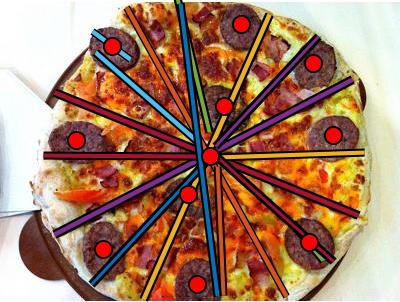}  &
		\includegraphics[width=\figGTWidth]{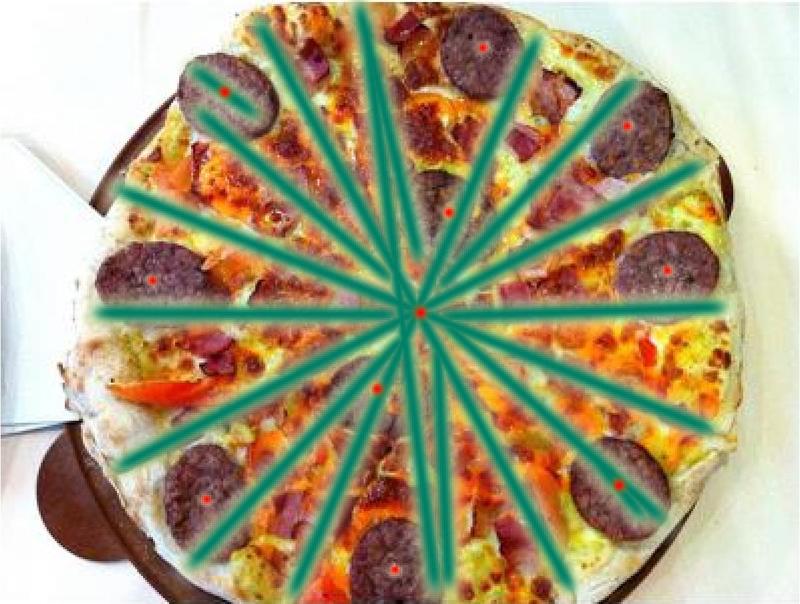} &
		\includegraphics[width=\figGTWidth]{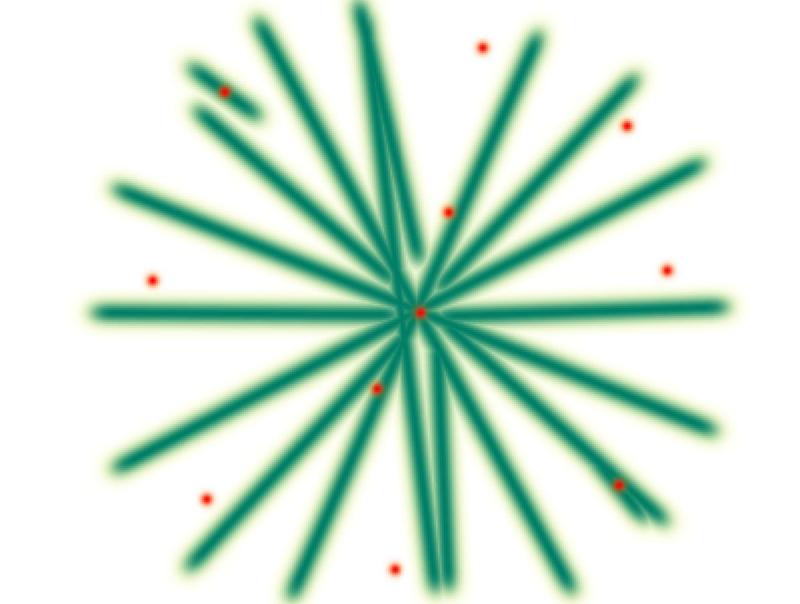} 
		\\
		\includegraphics[width=\figGTWidth]{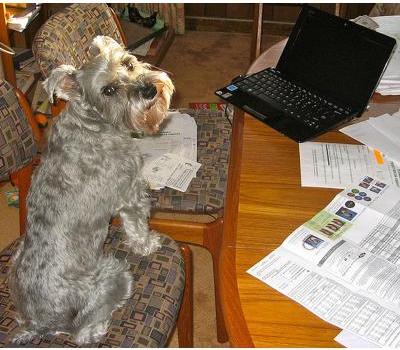} &
		\includegraphics[width=\figGTWidth]{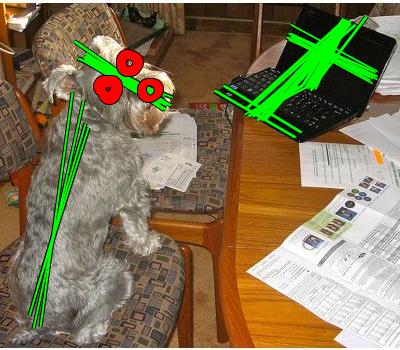} &
		\includegraphics[width=\figGTWidth]{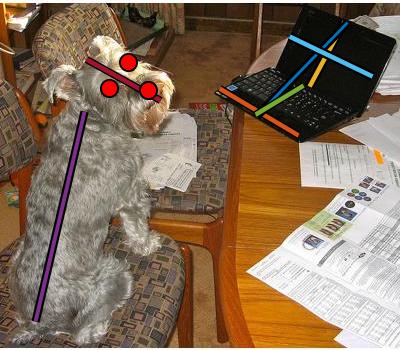}  &
		\includegraphics[width=\figGTWidth]{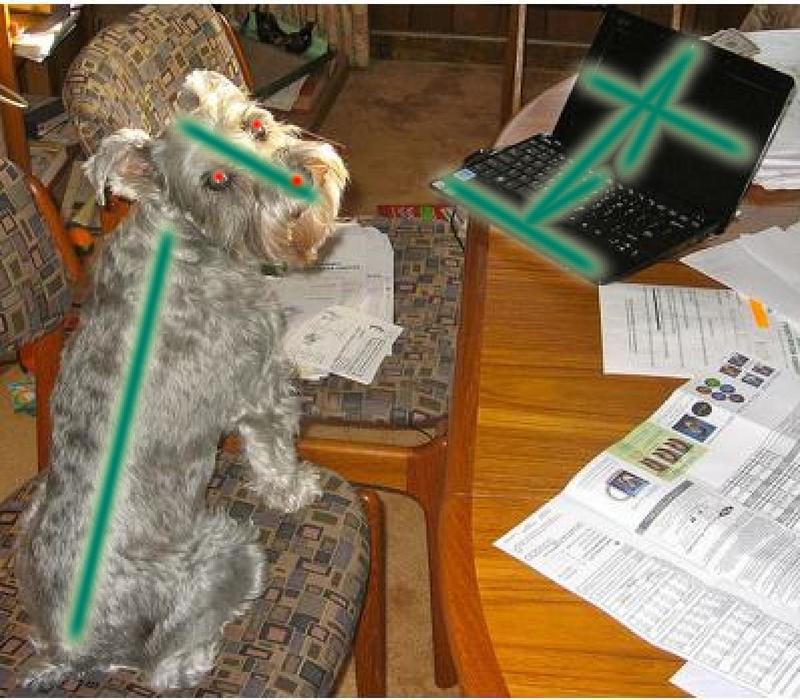} &
		\includegraphics[width=\figGTWidth]{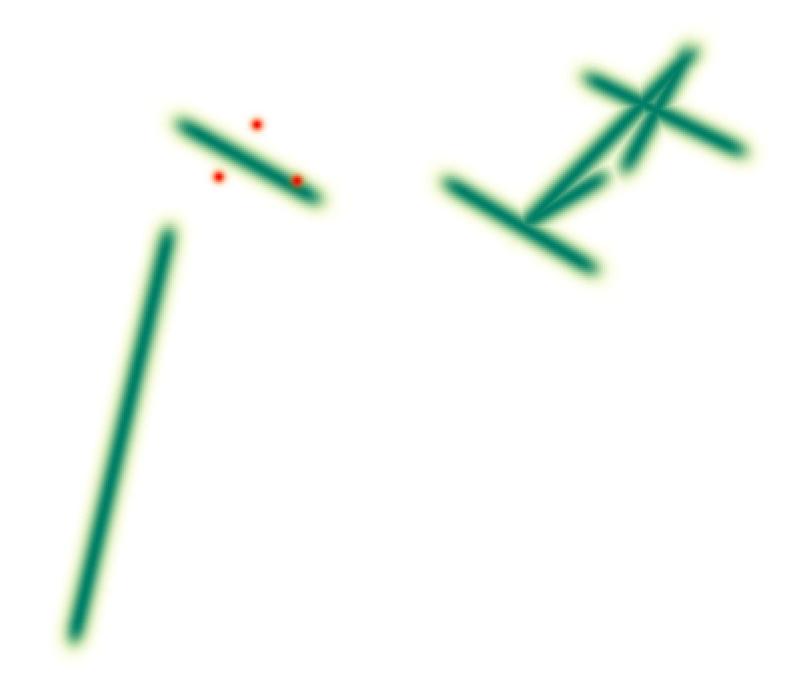} \\
	\end{tabular}
	\setlength{\tabcolsep}{2pt} 
	\caption{The progression (left to right) of converting the human labeled ground truth symmetries into symmetry heatmaps ($ H $).  The human labels are clustered to find the reflection symmetry axes and rotation symmetry centers.  Reflection symmetry heatmap: \textcolor{green}{green}, rotation: \textcolor{red}{red}.  } \label{fig:creating ground-truth} 
	 \vspace{-10pt}
\end{figure}

This is done by drawing a point for rotation center or a line for reflection axis on an image initialized with 0's and convolving with a Gaussian filter.  The resulting GT heatmap is then scaled between [0,1].  The max is taken among all individual GT heatmaps in an image so that nearby labels or intersecting lines do not create artifacts in the heatmap, similar to \cite{cao2016realtime}:
\begin{equation}
H_{i,j} = \max_k H_{i,j,k}\ .
\end{equation}
This assures that the heatmap is 1.0 at each rotation center and reflection axis and decreases exponentially as it moves away.  Sample heatmaps generated from human labels are shown in Figure \ref{fig:creating ground-truth}.
The GT images are augmented by random operations, including cropping, scaling ([.85,.9,1.1,1.25]), rotating ([0$ ^\circ $,90$ ^\circ $,180$ ^\circ $,270$ ^\circ $]), and reflecting w.r.t. the vertical central axis of the image. 

\subsection{Network}
We use two different networks, Sym-VGG is based on VGG-16~\cite{simonyan2014very} and Sym-ResNet on ResNet-101~\cite{he2016deep}.  Sym-VGG uses a similar structure to VGG-16 network with a dilation of 2 pixels at the conv5 layer and then an atrous pyramid~\cite{chen2016deeplab}.  Sym-ResNet has multi-scale streams (for the entire network) of 50\%, 75\%, and 100\% of the original image scale and dilation at the later layers.  Each scale has a separate atrous convolution pyramid and is fused together using a max operation similar to \cite{chen2016deeplab}. 
The final layer of the network upsamples the output using bilinear interpolation to the actual image size (8x) and then a $ \ell_2 $ loss is computed.  This upsampling eliminates any artifacts created from downsampling the ground-truth labels and trains the network to adapt to the upsampling.  
We use a similar strategy of borrowing weights from networks previously trained on Imagenet and fine-tuned on MS-COCO~\cite{chen2016deeplab,long2015fully,yosinski2014transferable}.  
This design strategy has been shown to be useful for image segmentation \cite{chen2014semantic,chen2016deeplab,long2015fully}, and allows us to train much larger networks without the need for millions of images. 
Atrous convolution~\cite{chen2014semantic,chen2016deeplab,holschneider1990real,papandreou2015modeling} is useful to provide contextual information for each pixel.  The context around each point proves to be crucial since  symmetry detection is about finding relationships between pixels (parts).  

\begin{figure*}[!t]
	\centering
	
	\newcounter{imageNumber}
	\setcounter{imageNumber}{1}
	\newlength{\imageRowWidthRef}
	\setlength{\imageRowWidthRef}{.111\linewidth} 
	\newcommand{\imgRowSingleRef}[2]{
		\parbox{#2\imageRowWidthRef}{
			%
			{\Large \sffamily \textbf{\Alph{imageNumber}}}
			\stepcounter{imageNumber}
			\vspace{-35pt} 
			\begin{center}
				\hspace*{2.5ex}\includegraphics[width=1\linewidth]{./results_images/Reflection/im/#1}
			\end{center}
			\vspace{-11.25pt} 
		} &
		\parbox{#2\imageRowWidthRef}{\centering \includegraphics[width=1\linewidth]{./results_images/Reflection/gt2/#1}} &
		\parbox{#2\imageRowWidthRef}{\centering \includegraphics[width=1\linewidth]{./results_images/Reflection/Loy/#1}} &
		\parbox{#2\imageRowWidthRef}{\centering \includegraphics[width=1\linewidth]{./results_images/Reflection/gray_jet_mil/#1}} &
		\parbox{#2\imageRowWidthRef}{\centering \includegraphics[width=1\linewidth]{./results_images/Reflection/gray_jet_srf/#1}} &
		\parbox{#2\imageRowWidthRef}{\centering \includegraphics[width=1\linewidth]{./results_images/Reflection/gray_jet_fsds/#1}} &
		\parbox{#2\imageRowWidthRef}{\centering \includegraphics[width=1\linewidth]{./results_images/Reflection/gray_jet_DeepLabv2_VGG16/#1}} &
		\parbox{#2\imageRowWidthRef}{\centering \includegraphics[width=1\linewidth]{./results_images/Reflection/gray_jet_DeepLabv2_ResNet101/#1}} \\}

	\renewcommand{\arraystretch}{.2}
	\setlength{\tabcolsep}{2pt}
	\begin{tabular}{*{8}{c}}
		\multicolumn{8}{c}{\textbf{\large Reflection Symmetry Detection}} \\
		&&&&&&\multicolumn{2}{c}{Proposed Sym-NETs}\\
		\hspace*{5ex}\parbox{\imageRowWidthRef}{\centering Original Image} & 
		\parbox{\imageRowWidthRef}{\centering Ground Truth} &
		\parbox{\imageRowWidthRef}{\centering Loy \& Eklundh~\cite{loy2006detecting}} &
		\parbox{\imageRowWidthRef}{\centering MIL~\cite{tsogkas2012learning}} &
		\parbox{\imageRowWidthRef}{\centering SRF~\cite{teo2015detection}} & 
		\parbox{\imageRowWidthRef}{\centering FSDS~\cite{shen2016object}} &
		\parbox{\imageRowWidthRef}{\centering Sym-VGG}   &
		\parbox{\imageRowWidthRef}{\centering Sym-ResNet}  \\
		\imgRowSingleRef{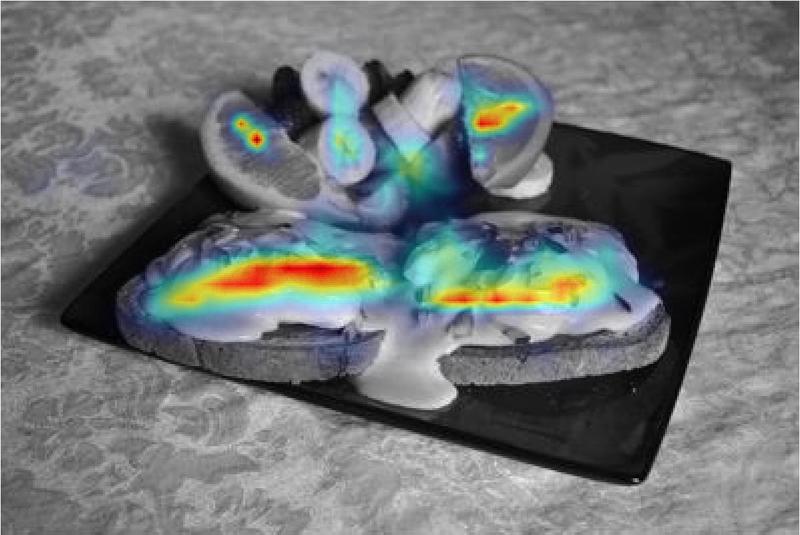}{1} \\
		\imgRowSingleRef{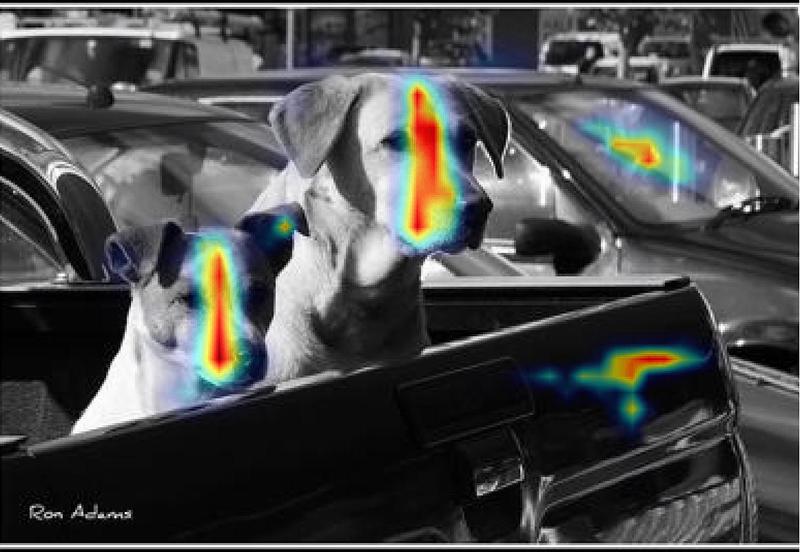}{1} \\
		\imgRowSingleRef{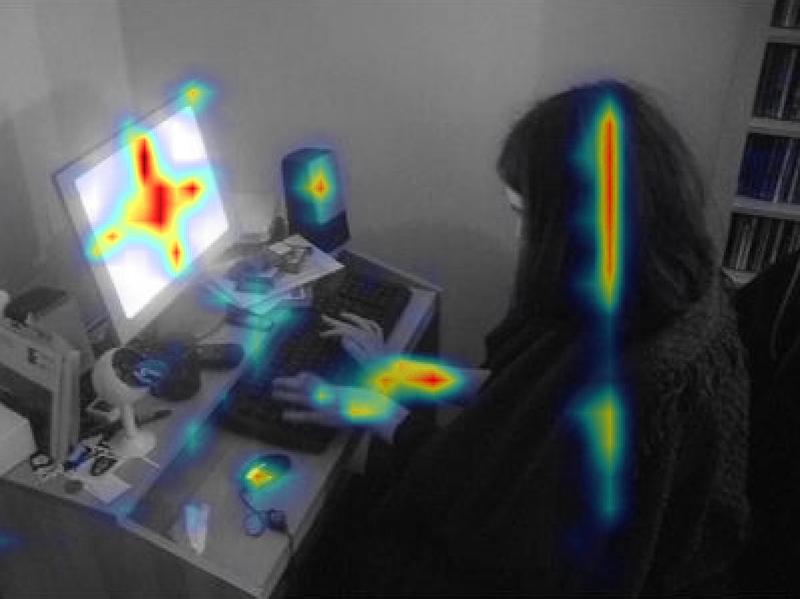}{1} \\
		\imgRowSingleRef{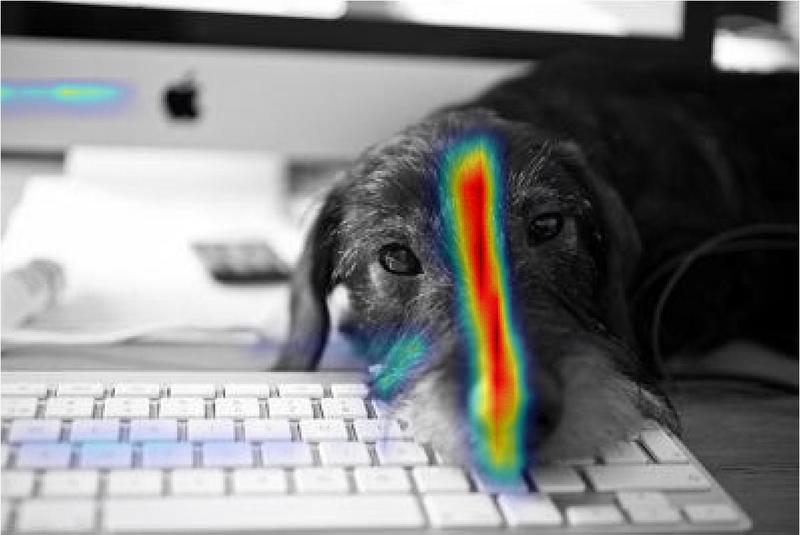}{1} \\	
		\imgRowSingleRef{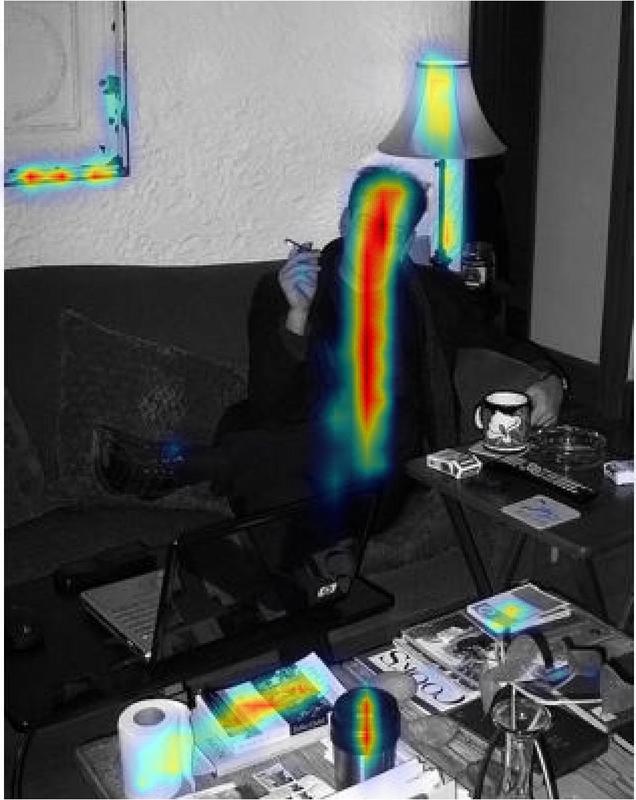}{1} \\
		\imgRowSingleRef{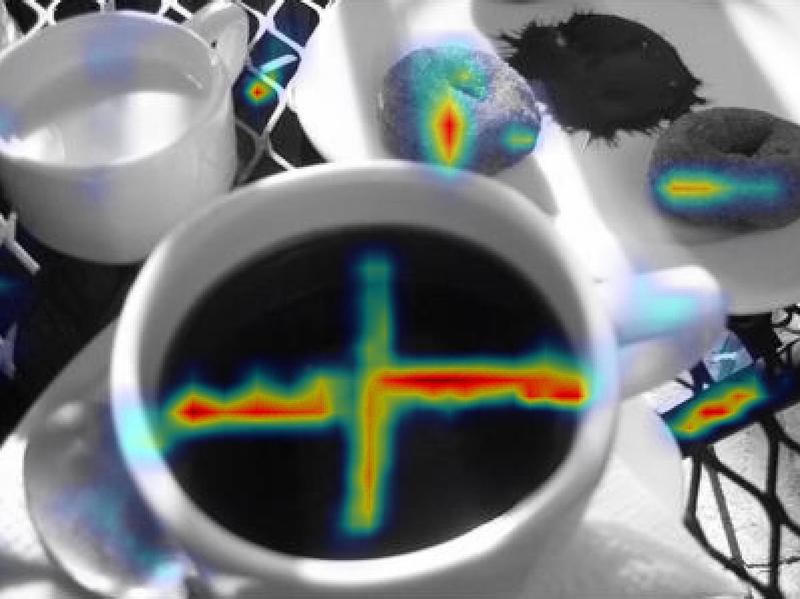}{1} \\
%
	\end{tabular}
	
	\caption{Examples of the original image, ground-truth, and the output for the reflection symmetry detection algorithms before thinning. }
	\label{fig:Montage of Images ref} \vspace{-10pt}
\end{figure*}

\begin{figure*}[t!]
	\centering
	
	\setcounter{imageNumber}{1}
	\newlength{\imageRowWidthRot}
	\setlength{\imageRowWidthRot}{.1125\linewidth}
	\newcommand{\imgRowSingleRot}[2]{
		\parbox{#2\imageRowWidthRot}{
			%
			{\Large \sffamily \textbf{\Alph{imageNumber}}}
			\stepcounter{imageNumber}
			\vspace{-35pt} 
			\begin{center}
				\hspace*{2.5ex}\includegraphics[width=1\linewidth]{./results_images/Rotation/im/#1}
			\end{center}
			\vspace{-11.25pt} 
		} &
		\parbox{#2\imageRowWidthRot}{\centering \includegraphics[width=1\linewidth]{./results_images/Rotation/gt2/#1}} &
		\parbox{#2\imageRowWidthRot}{\centering \includegraphics[width=1\linewidth]{./results_images/Rotation/Loy/#1}} &
		\parbox{#2\imageRowWidthRot}{\centering \includegraphics[width=1\linewidth]{./results_images/Rotation/gray_jet_DeepLabv2_VGG16/#1}} &
		\parbox{#2\imageRowWidthRot}{\centering \includegraphics[width=1\linewidth]{./results_images/Rotation/blank_DeepLabv2_VGG16/#1}} &
		\parbox{#2\imageRowWidthRot}{\centering \includegraphics[width=1\linewidth]{./results_images/Rotation/gray_jet_DeepLabv2_ResNet101/#1}} &
		\parbox{#2\imageRowWidthRot}{\centering \includegraphics[width=1\linewidth]{./results_images/Rotation/blank_DeepLabv2_ResNet101/#1}} \\}
	
	\setlength{\tabcolsep}{3pt}
	\renewcommand{\arraystretch}{.2}
	\begin{tabular}{*{7}{c}}
		\multicolumn{7}{c}{\textbf{\large Rotation Symmetry Detection}} \\
		&&&\multicolumn{4}{c}{Proposed Sym-NETs}\\
		\hspace*{5ex}\parbox{\imageRowWidthRot}{\centering Original Image} & 
		\parbox{\imageRowWidthRot}{\centering Ground Truth} &
		\parbox{\imageRowWidthRot}{\centering Loy \& Eklundh~\cite{loy2006detecting}}   &
		\multicolumn{2}{c}{\parbox{\imageRowWidthRot}{\centering Sym-VGG}}   &
		\multicolumn{2}{c}{\parbox{\imageRowWidthRot}{\centering Sym-ResNet}}  \\
		\imgRowSingleRot{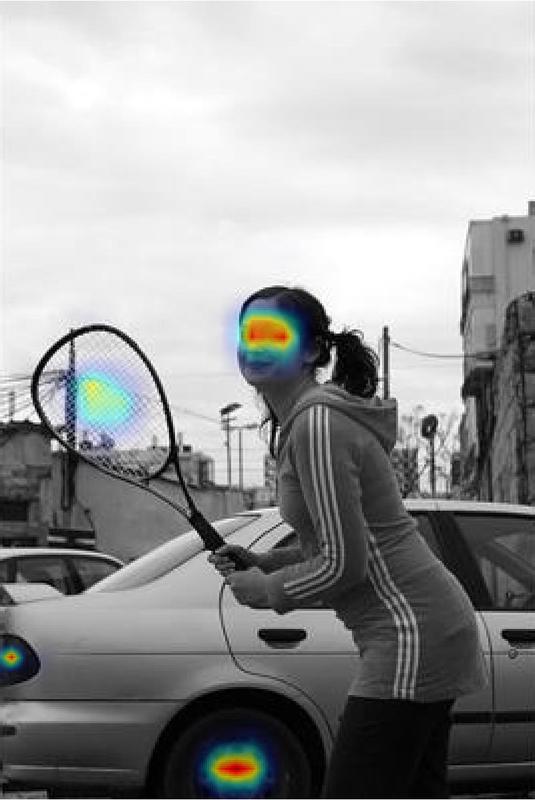}{.7} \\
		\imgRowSingleRot{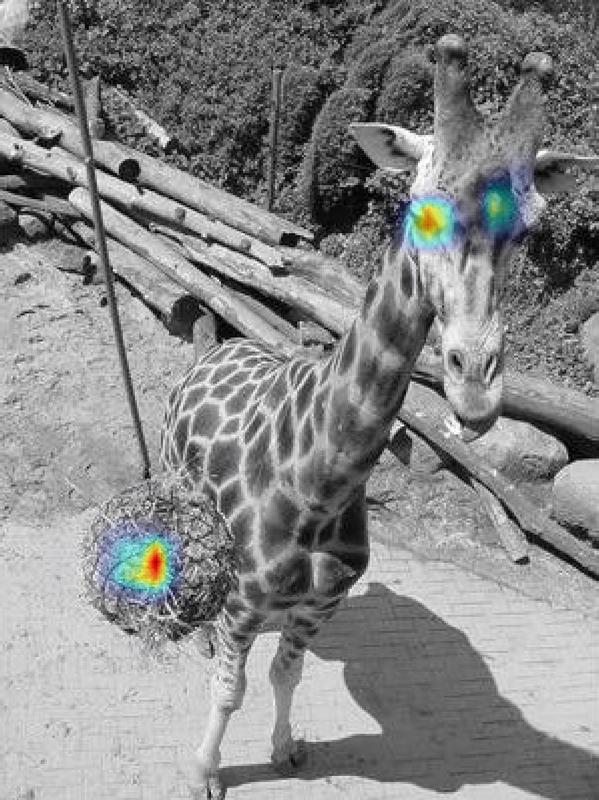}{.7} \\
		\imgRowSingleRot{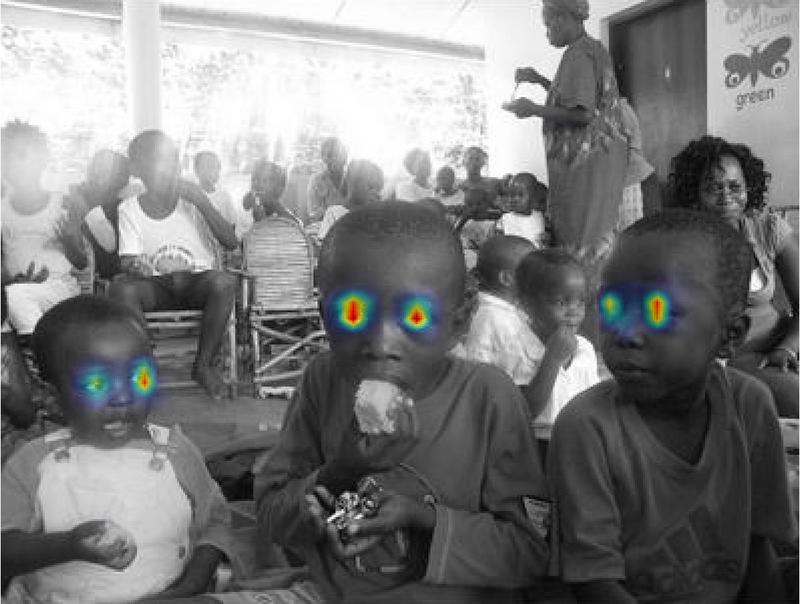}{1} \\
		\imgRowSingleRot{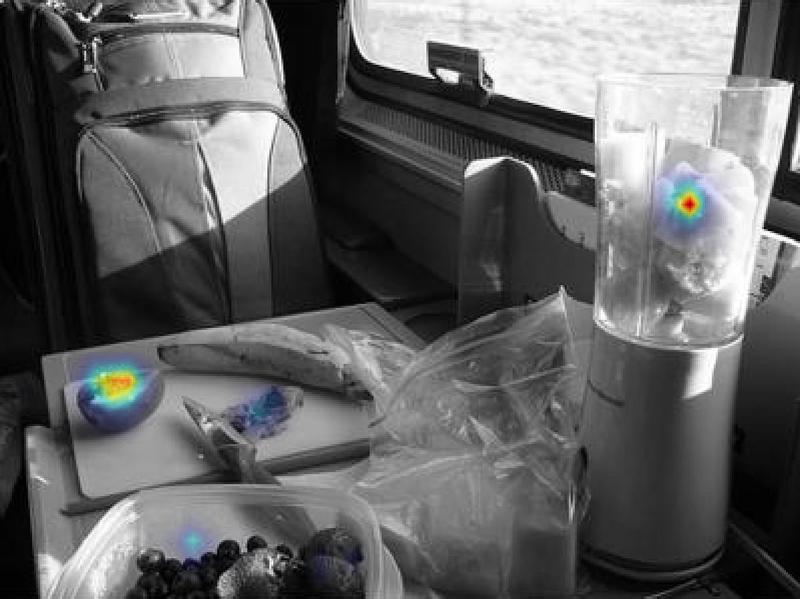}{1} \\
		\imgRowSingleRot{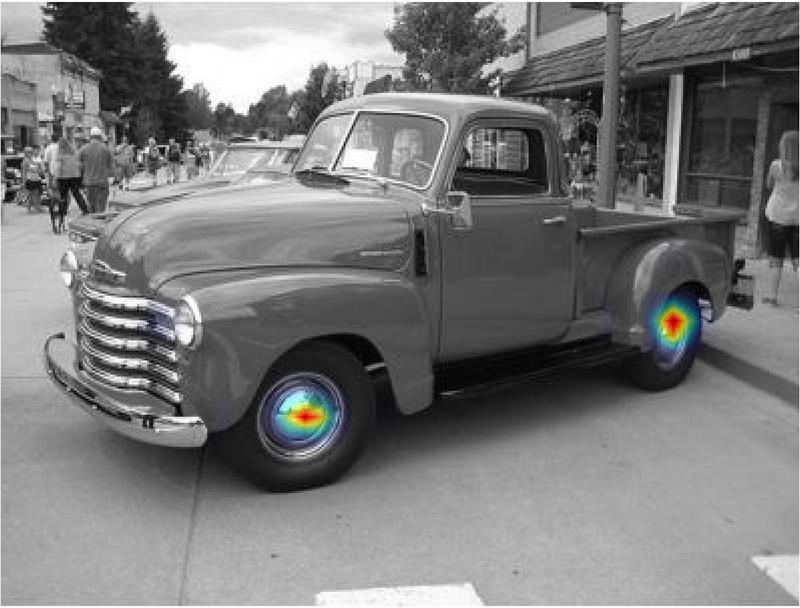}{1} \\
		\imgRowSingleRot{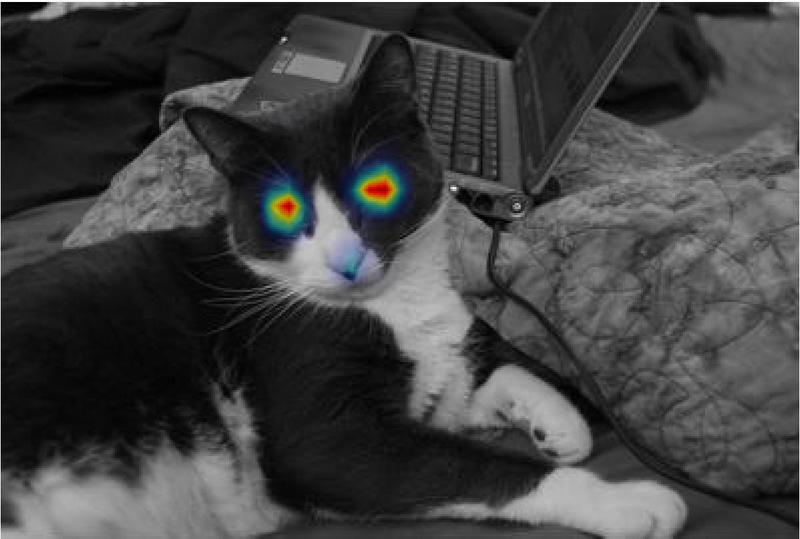}{1} \\
		\imgRowSingleRot{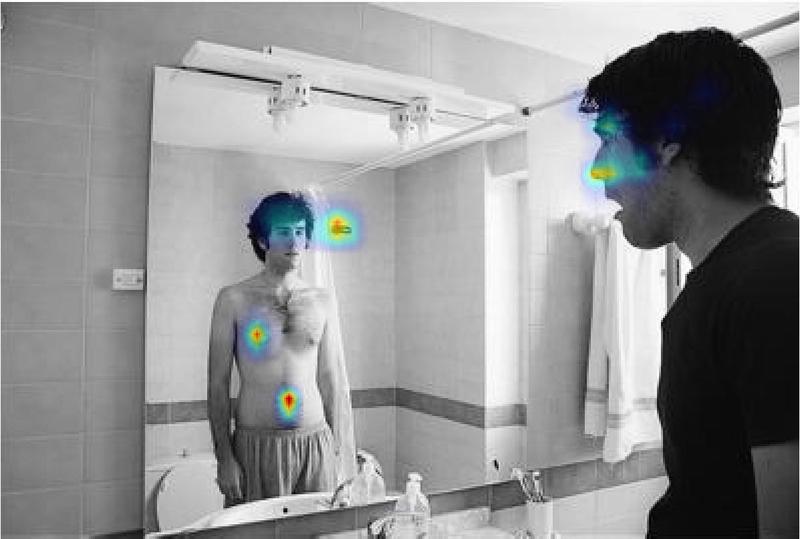}{1} \\
%
	\end{tabular} \vspace{2pt}
	\caption{Examples of the original image, ground-truth, and the output for the rotation detection algorithms.  Rotation symmetry heatmaps are shown for both types of Sym-NETs.  }
	\label{fig:Montage of Images rot} \vspace{-10pt}
\end{figure*}
\setlength{\tabcolsep}{6pt}

\subsection{Training}
\label{sec:Training}
We train Sym-VGG and Sym-ResNet separately for reflection and rotation symmetry detection.  
We use an 80\%/20\% split of 1202 images from the MS-COCO dataset~\cite{lin2014microsoft} with at least one GT for each type.  The dataset includes 1199 and 1057 images with reflection and rotation symmetry ground-truth. This creates train/test datasets of 959/240 images for reflection and 846/211 for rotation.  Each network is trained with an exponential learning rate multiplier of $ 1-\frac{batch\;number}{total\;batches}^{power} $ similar to other recent segmentation networks~\cite{chen2016deeplab} in the Caffe framework~\cite{jia2014caffe}. 

The reflection Sym-NETs use a learning rate of 1e-10 and 2.5e-11 and the rotation Sym-NETs use a rate of 1e-9 and 2.5e-10 for the VGG and Resnet networks respectively.  The learning rates are empirically found.  The Sym-VGG takes 3 days and the Sym-ResNet takes 10 days to converge on a GeForce GTX Titan~X.


\begin{figure*}[!t]
	\centering
	\newlength{\dataWidth}
	\newlength{\labelWidth}
	\setlength{\dataWidth}{.45\linewidth}
	\setlength{\labelWidth}{.48\linewidth}
	\begin{tabular}{c|c} 
		\parbox{.45\linewidth}{\centering \textbf{Testset from MS-COCO~\cite{lin2014microsoft}}} &
		\parbox{.45\linewidth}{\centering \textbf{Testset from CVPR'13 Symmetry~Competition~\cite{liu2013symmetry}}} \\
		\parbox{\labelWidth}{{\large \sffamily \textbf{A}} \vspace{-30pt}} &
		\parbox{\labelWidth}{{\Large \sffamily \textbf{C}} \vspace{-30pt}} \\
		\parbox{\dataWidth}{\raggedleft\includegraphics[width=\linewidth]{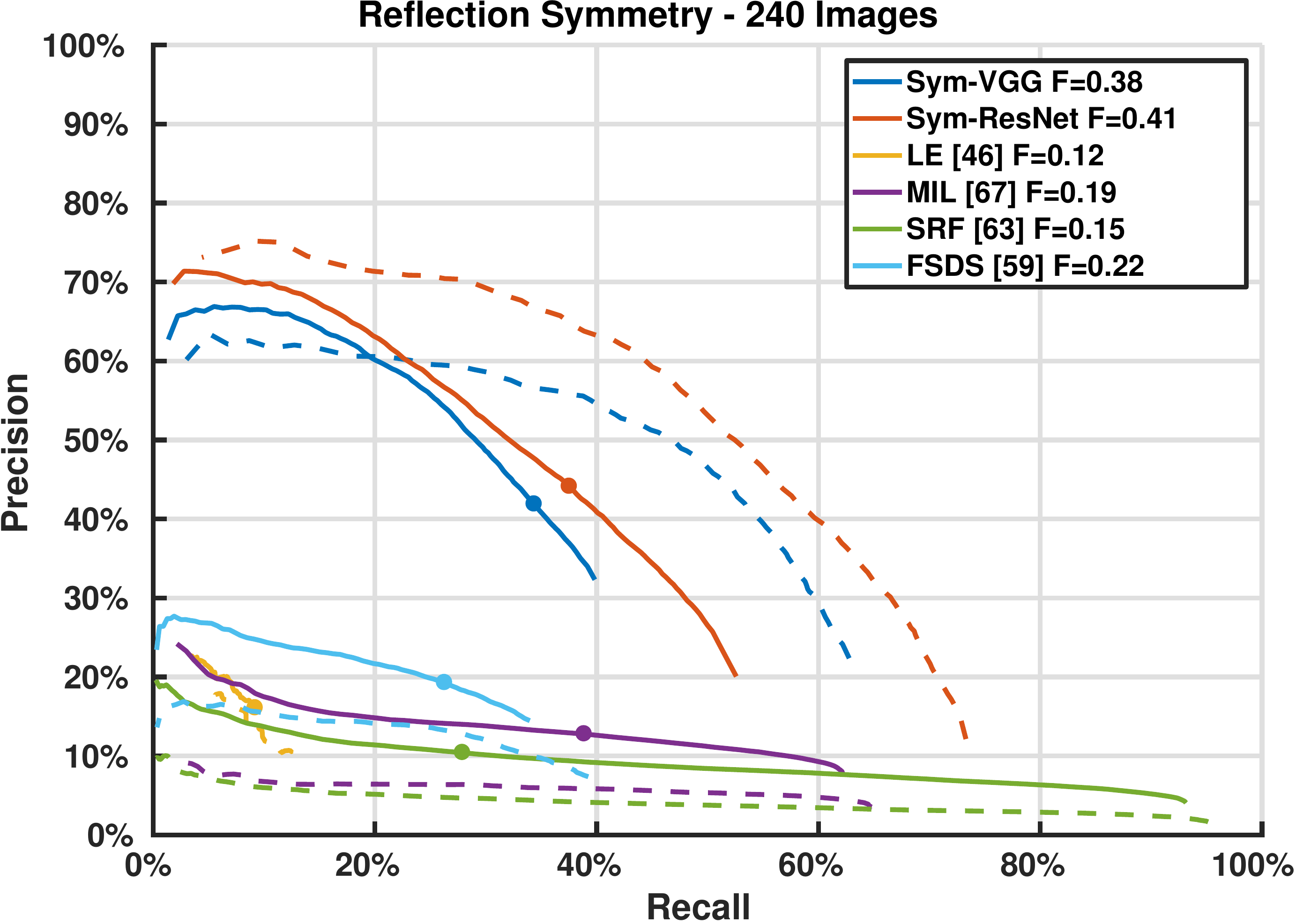}} &
		\parbox{\dataWidth}{\raggedleft\includegraphics[width=\linewidth]{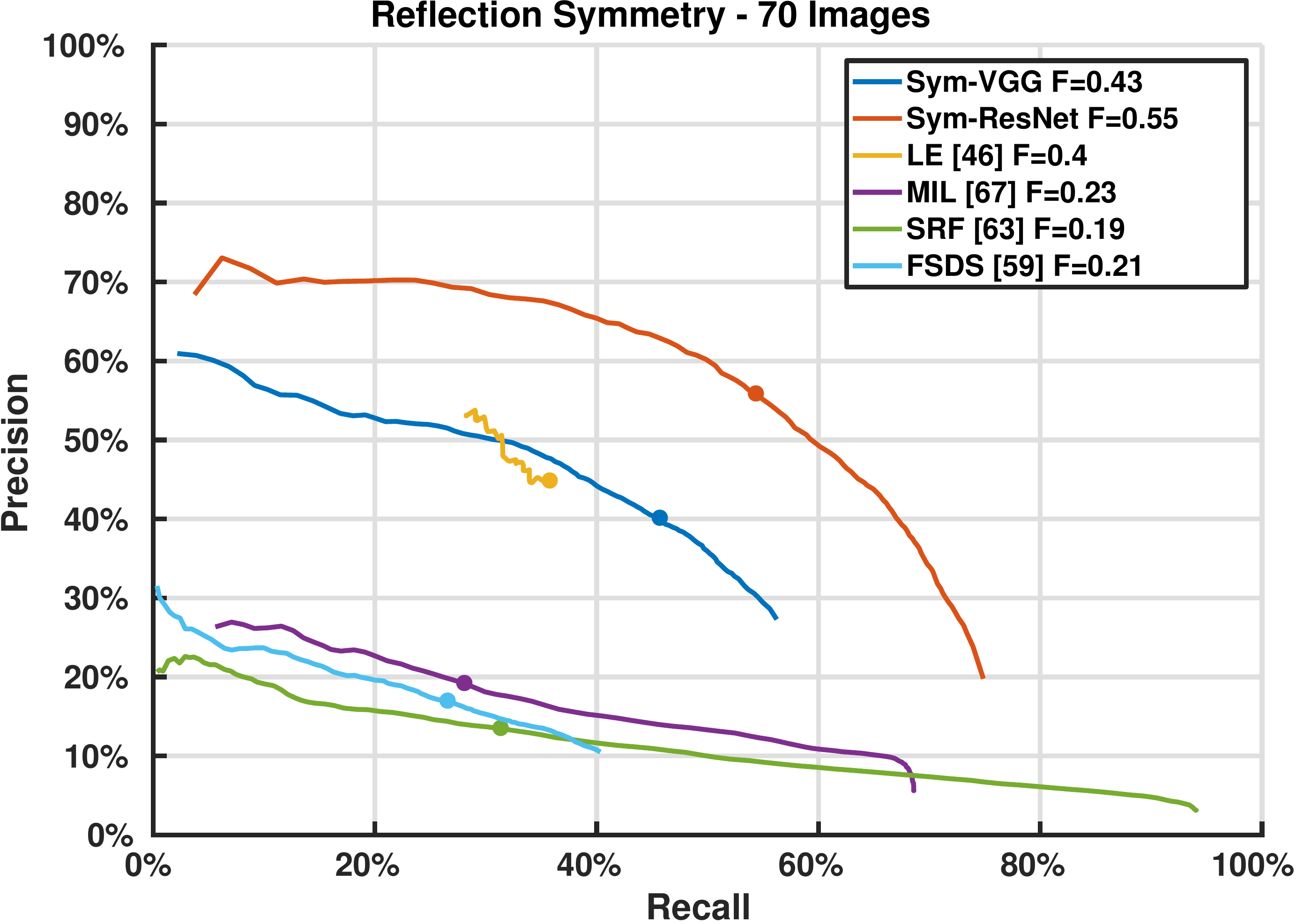}} \\
		\parbox{\labelWidth}{{\Large \sffamily \textbf{B}} \vspace{-30pt}} &
		\parbox{\labelWidth}{{\Large \sffamily \textbf{D}} \vspace{-30pt}} \\
		\parbox{\dataWidth}{\raggedleft\includegraphics[width=\linewidth]{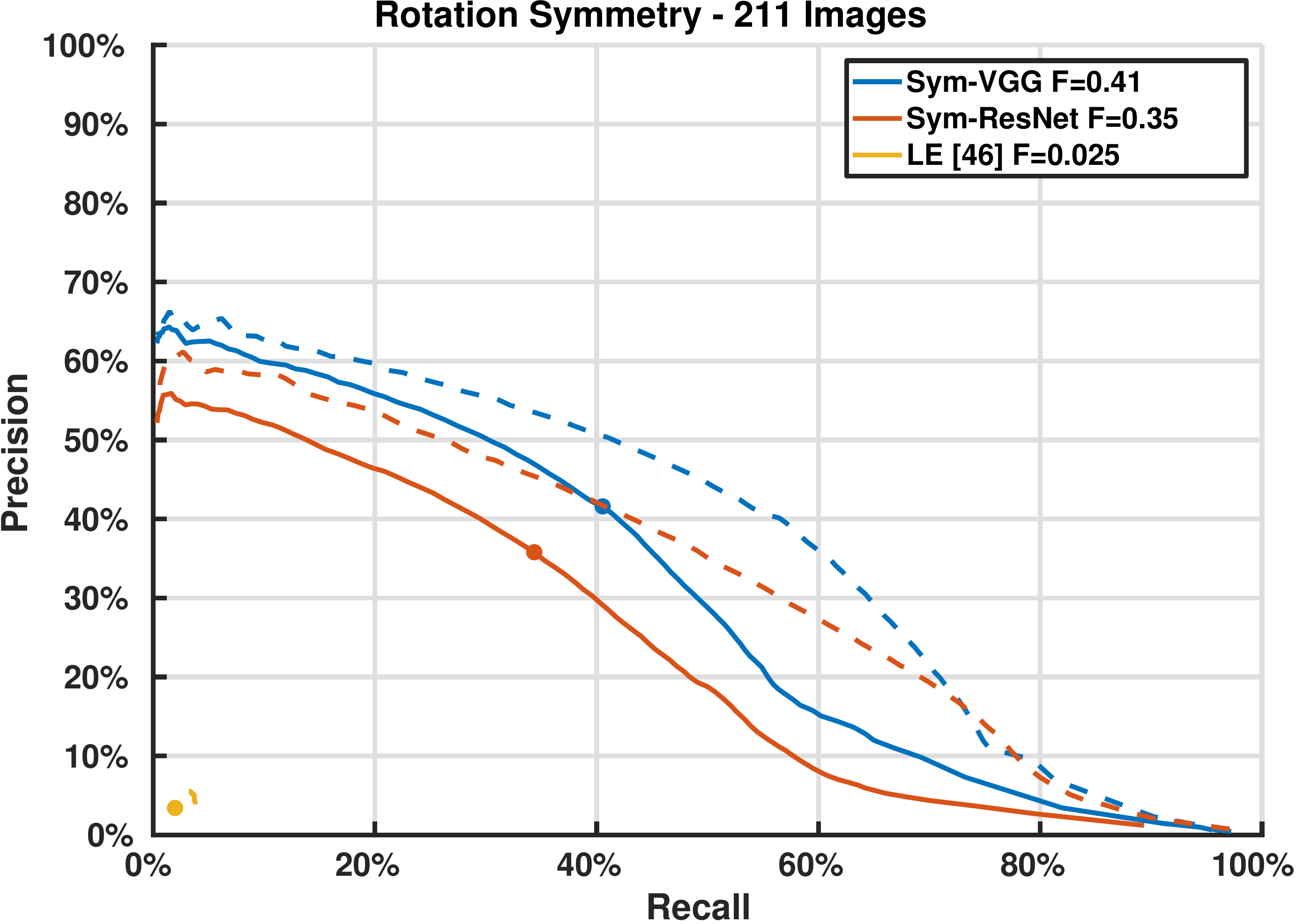}}  &
		\parbox{\dataWidth}{\raggedleft\includegraphics[width=\linewidth]{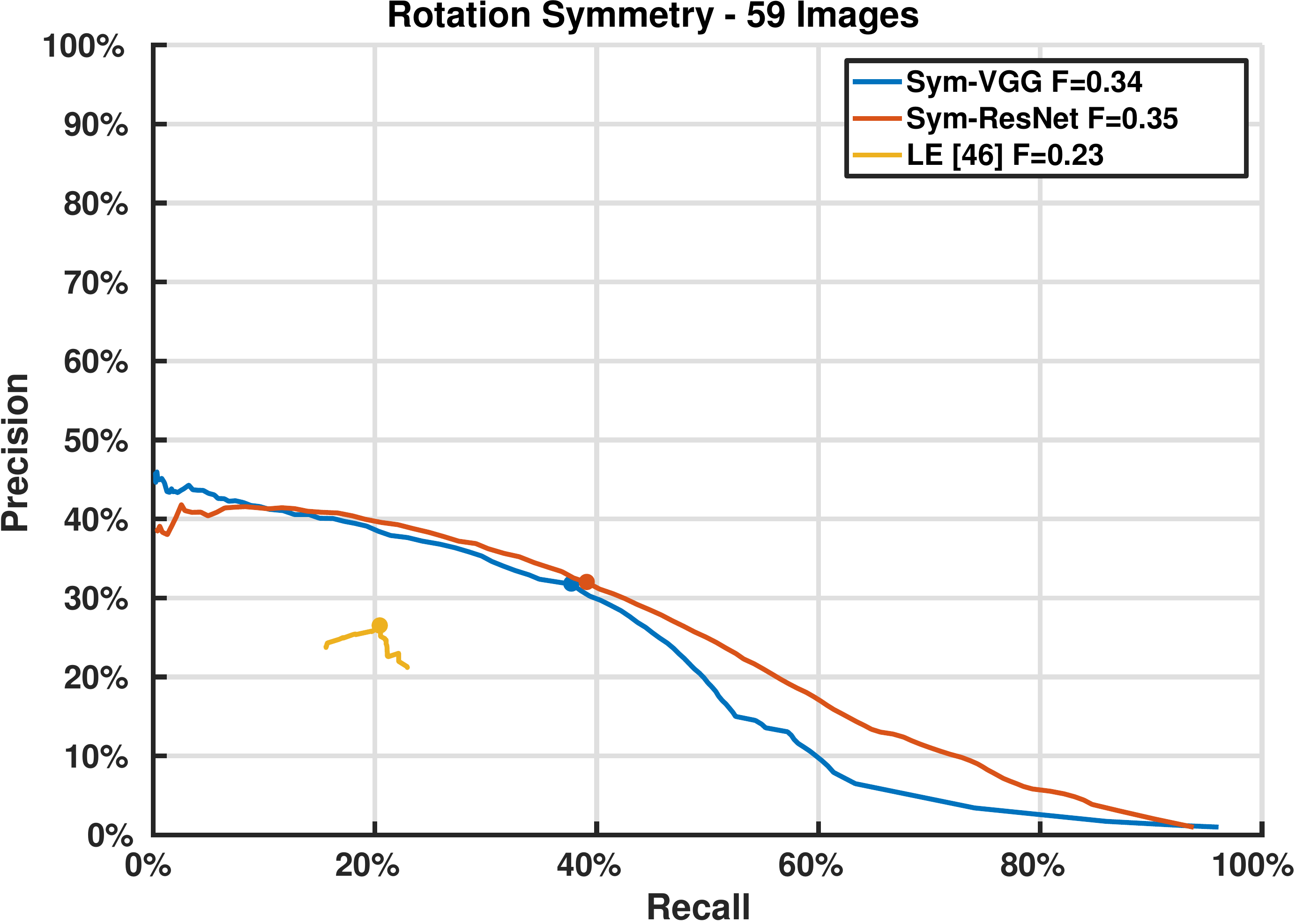}}  \\
	\end{tabular} \vspace{5pt}
	\caption{Comparison of the precision-recall curves for state-of-the-art symmetry detection algorithms.  \textbf{A and B:} The comparison on MS-COCO images for all GT labels (solid line) and for the subset of 184 images with GT labels containing at least 20 labelers (dashed line), and the maximum F-measure values (dot on the line). 
	\textbf{C and D:} comparison on the test image set of the  CVPR 2013 symmetry competition~\cite{liu2013symmetry}.  
	Best viewed electronically. }
	\label{fig:PR_ref} \vspace{-12.5pt}
\end{figure*}

\subsection{Performance Evaluation}
\label{sec:evaluation}
Figures~\ref{fig:Montage of Images ref} and \ref{fig:Montage of Images rot} show sample 
reflection and rotation detection results.  
To quantitatively evaluate the performance of the networks, we compute a 
precision-recall curve for each symmetry detector in a similar way to \cite{martin2004learning,tsogkas2012learning}, which is
generated by stepping through 100 thresholds (between~[0,1]) on the networks' heatmap output.
From these scores, we also calculate the maximum F-measures $  ( 2\times \frac{\text{precision } + \text{ recall}}{\text{precision } \times \text{ recall}} ) $~\cite{martin2004learning,shen2016object,tsogkas2012learning} for each symmetry detector to obtain a single value as an indicator of its statistical strength~\cite{martin2004learning,shen2016object,tsogkas2012learning}.  For reflection, we use a one pixel-width reflection axis as the ground-truth  \cite{martin2004learning,shen2016object,tsogkas2012learning} 
and use the measure defined in \cite{martin2004learning,tsogkas2012learning}. 
%
For rotation, we use a 5-pixel radius ($\tau$) circle around the GT symmetry \cite{funk_liuCVPR2016Symmetry} and calculate the explicit overlap.



%

%
\subsection{Performance Comparison}

Not only would we like to know which algorithm performs better on a given test set, we would also like to 
demonstrate whether the better performance is statistically significant. 
In this comparison study, we use the maximum F-Measure computed from its mean precision-recall rate (Section \ref{sec:evaluation}) in order to compare all detectors at their respective optimal values.  We then use a paired t-test on max F-measures between pairs of symmetry detectors and obtain the p-value indicating the significance level of their difference. 

We compare the output of our symmetry detection system with both dense and sparse symmetry detection algorithms qualitatively (Figures~\ref{fig:Montage of Images ref} and \ref{fig:Montage of Images rot}) and quantitatively (Figure~\ref{fig:PR_ref}).  For sparse detection, we use Loy and Eklundh's~(\textit{LE})~\cite{loy2006detecting} algorithm, a simple and fast SIFT-feature based reflection/rotation symmetry detector.  The sparse output from the algorithm is transformed into dense labels by applying the same operations that create the evaluation ground truth from their sparse labels, weighted by the algorithm's detection strength for each symmetry.


For dense detection algorithms, we include Tsogkas and Kokkinos' Multiple Instance Learning method~(\textit{MIL})~\cite{tsogkas2012learning}, Teo~\etal's~ method~(\textit{SRF})~\cite{teo2015detection}, and Deep Skeletonization network~(\textit{FSDS})~\cite{shen2016object} as a part of our comparison.
Our goal is to determine the performance difference between the {\em skeletonization} and reflection symmetry detection algorithms.  Even though there is a conceptual overlap on (local) symmetry, they do not detect the same types or ranges of symmetries.
The same non-maximum suppression algorithm \cite{dollar2015fast} is applied to the output of
Sym-NETs and the FSDS. All the default parameters for the algorithms are used in the comparison.  
On all datasets tested, at least one Sym-NET obtains significant improvement over the other detectors (Figure \ref{fig:PR_ref}).

\subsubsection{MS-COCO dataset}
We test the symmetry detectors against the MS-COCO~\cite{lin2014microsoft} dataset with symmetry labels 
(Section \ref{sec:Training}),  containing 240 reflection and 211 rotation images.
Both Sym-NETs significantly outperform all other detectors on the MS-COCO dataset for detecting the ground-truth symmetries derived from human labels (\mbox{P-value} $\ll 0.001$).

Furthermore, for the MS-COCO symmetry dataset, we take into account 
the number of labelers for each ground-truth symmetry. 
Symmetry GTs with less than 20 labels are taken out from this evaluation, creating subsets of 111 (of 240) reflection symmetry images and 73 (of 211) rotation symmetry images, representing the most prominent symmetries.  The statistics of the number of human labels for each symmetry is shown in Figure~\ref{fig:Max Number of Labeled Symmetries}.  
We observe that Sym-NETs perform better on detecting those symmetries perceived by humans as more prominent (more than 20 individual labelers for each symmetry) in the images (Figures~\hyperref[fig:PR_ref]{\ref{fig:PR_ref}A} and \hyperref[fig:PR_ref]{\ref{fig:PR_ref}B}).  

\subsubsection{CVPR 2013 Symmetry Competition Dataset}
Finally, 
to illustrate how Sym-NETs generalize onto other datasets, 
we use the test image sets from the CVPR 2013 symmetry competition~\cite{liu2013symmetry} 
with 70 reflection symmetry images and 59 rotation symmetry images.
%
Each image contains at least one labeled symmetry. 
During the past two CVPR symmetry detection competitions~\cite{liu2013symmetry,rauschert2011symmetry}, 
Loy and Eklundh's algorithm~\cite{loy2006detecting} 
has performed most competitively. Thus we compare Sym-NET output on CVPR test image sets against those of Loy and Eklundh~\cite{loy2006detecting}. These visual symmetries are relatively more well-defined on the image than the MS-COCO image set.

%
 All images and GTs of the CVPR 2013 testset are rescaled so the longest edge is at most 513 pixels (the maximum for our networks). 
The quantitative evaluations of the algorithm performance are shown in Figures~\hyperref[fig:PR_ref]{\ref{fig:PR_ref}C} and \hyperref[fig:PR_ref]{\ref{fig:PR_ref}D}.  
For both types of symmetries, Sym-ResNet remains significantly better than all algorithms evaluated, while the \mbox{F-measure} of Loy and Eklundh~\cite{loy2006detecting} is lower but on par with \mbox{Sym-VGG} statistically.
%

\begin{figure}[h]
	\centering
	\newlength{\imgWidth}
	\setlength{\imgWidth}{.18\linewidth}
	
	\setlength{\tabcolsep}{1.75pt}
	\begin{tabular}{ccccc}
		&
		\multicolumn{2}{c}{\parbox{\imgWidth}{\centering Reflection}} & 
		\multicolumn{2}{c}{\parbox{\imgWidth}{\centering Rotation}} \\
		\parbox{\imgWidth}{\centering Original Image} & 
		\parbox{\imgWidth}{\centering Sym-VGG} & 
		\parbox{\imgWidth}{\centering Sym-ResNet} &
		\parbox{\imgWidth}{\centering Sym-VGG} & 
	\parbox{\imgWidth}{\centering Sym-ResNet} \\
		\includegraphics[width=0.18\linewidth]{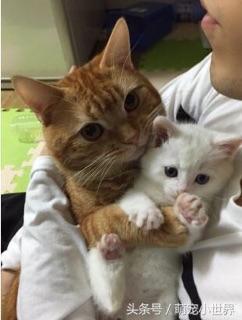} &
		\includegraphics[width=0.18\linewidth]{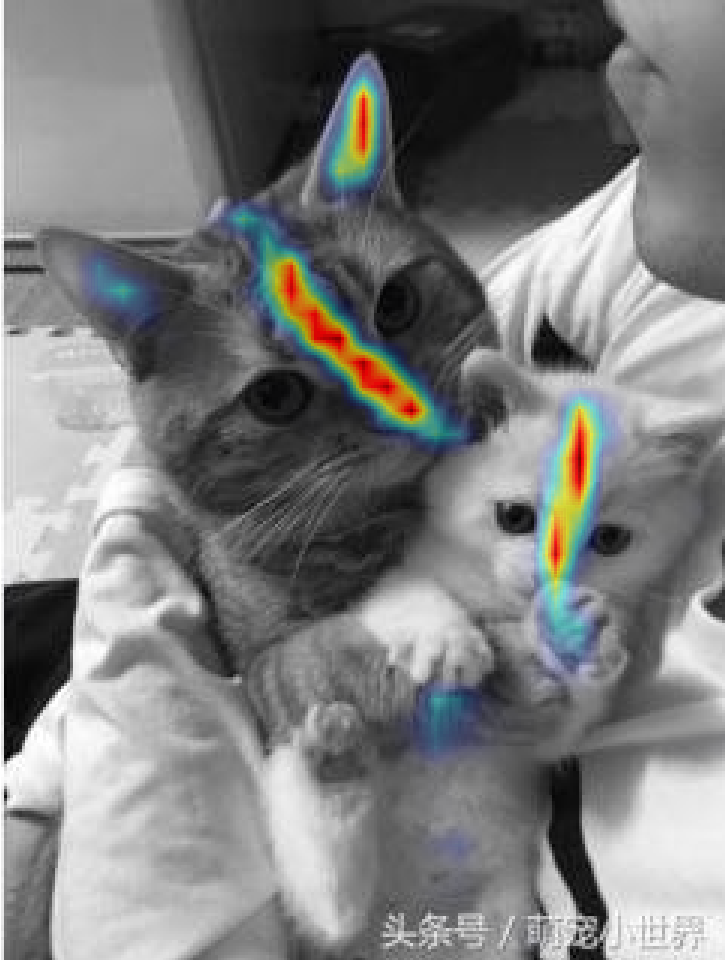} &
		\includegraphics[width=0.18\linewidth]{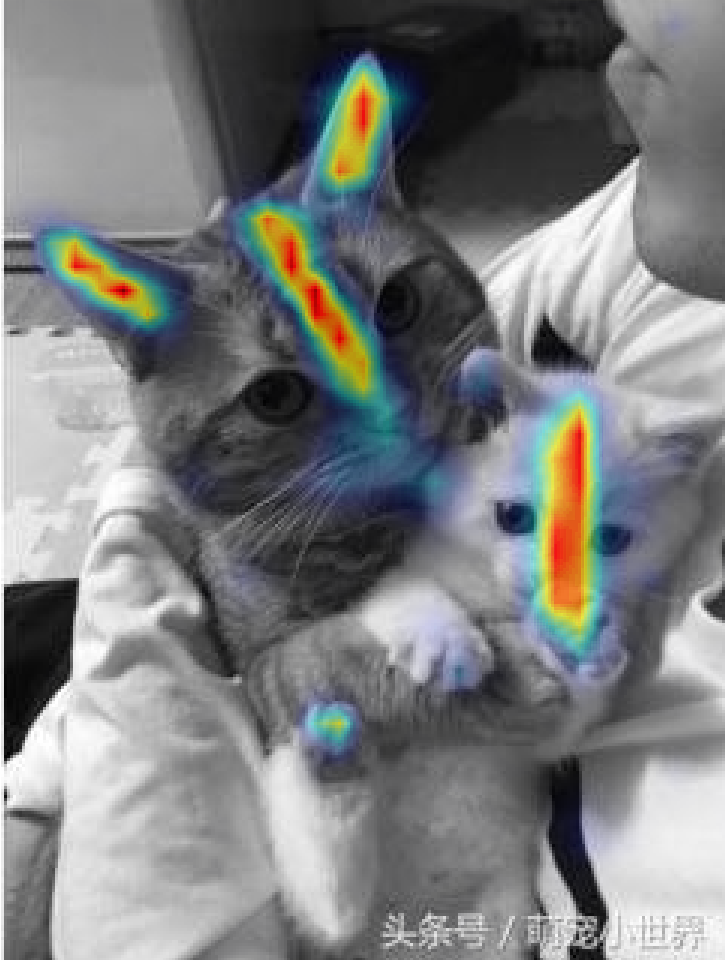} &
		\includegraphics[width=0.18\linewidth]{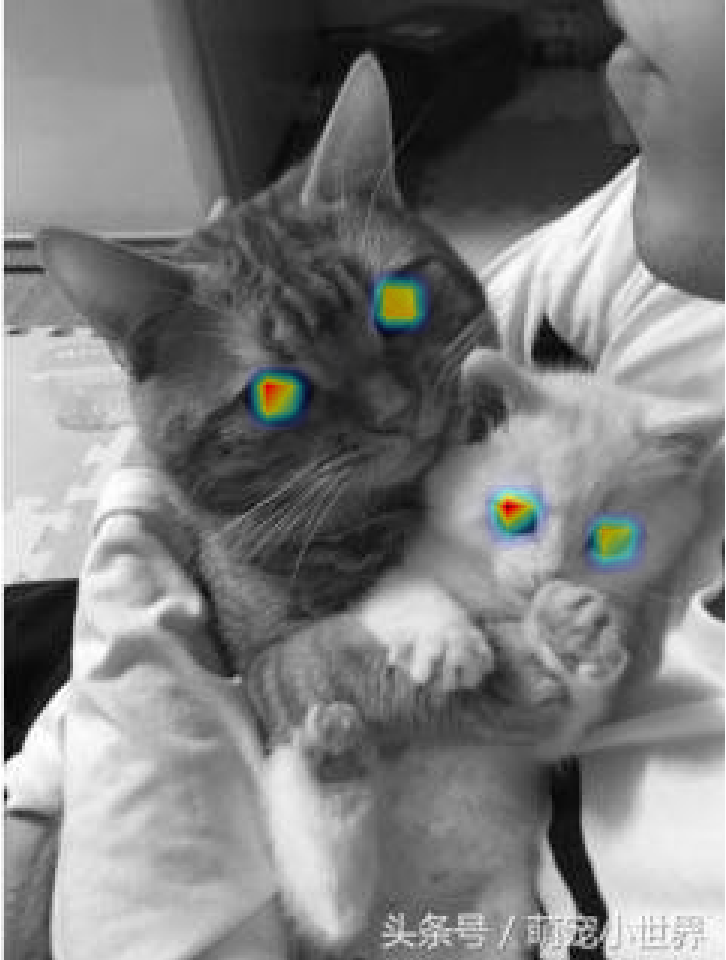} &
		\includegraphics[width=0.18\linewidth]{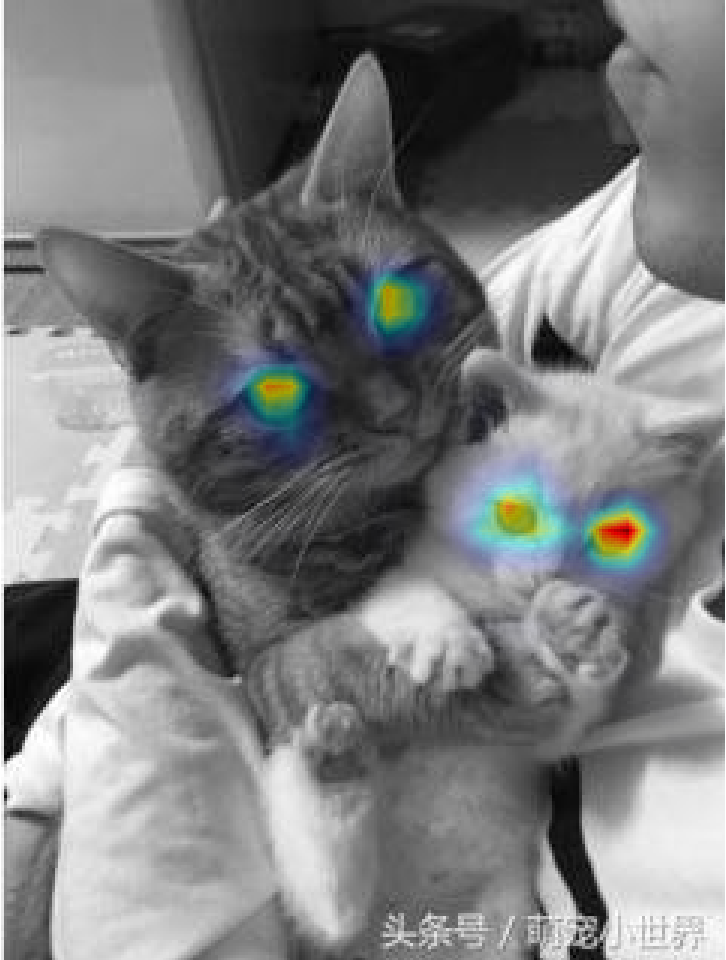} \\ 
		\includegraphics[width=0.18\linewidth]{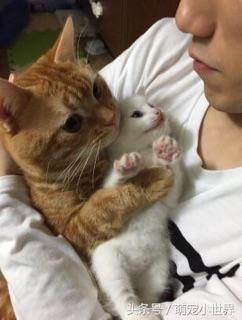} &
		\includegraphics[width=0.18\linewidth]{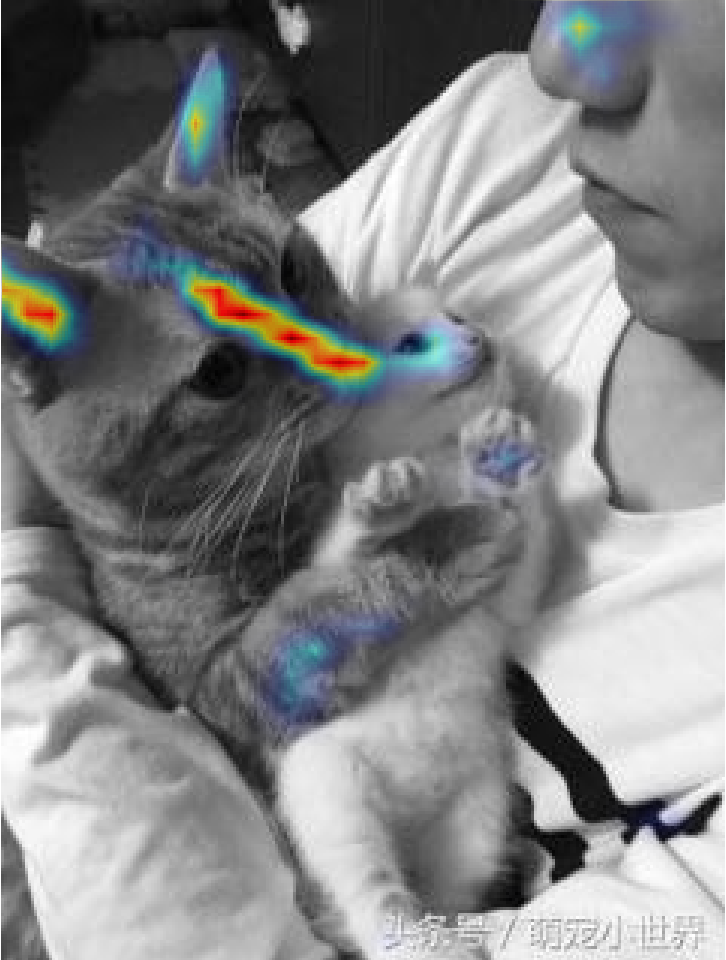} &
		\includegraphics[width=0.18\linewidth]{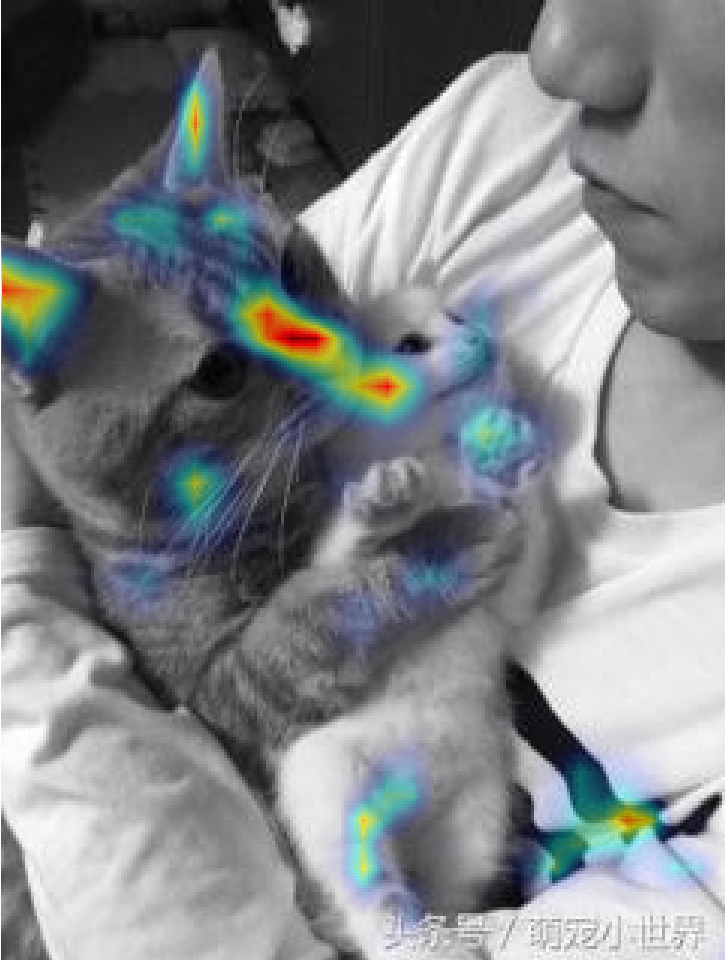} &
		\includegraphics[width=0.18\linewidth]{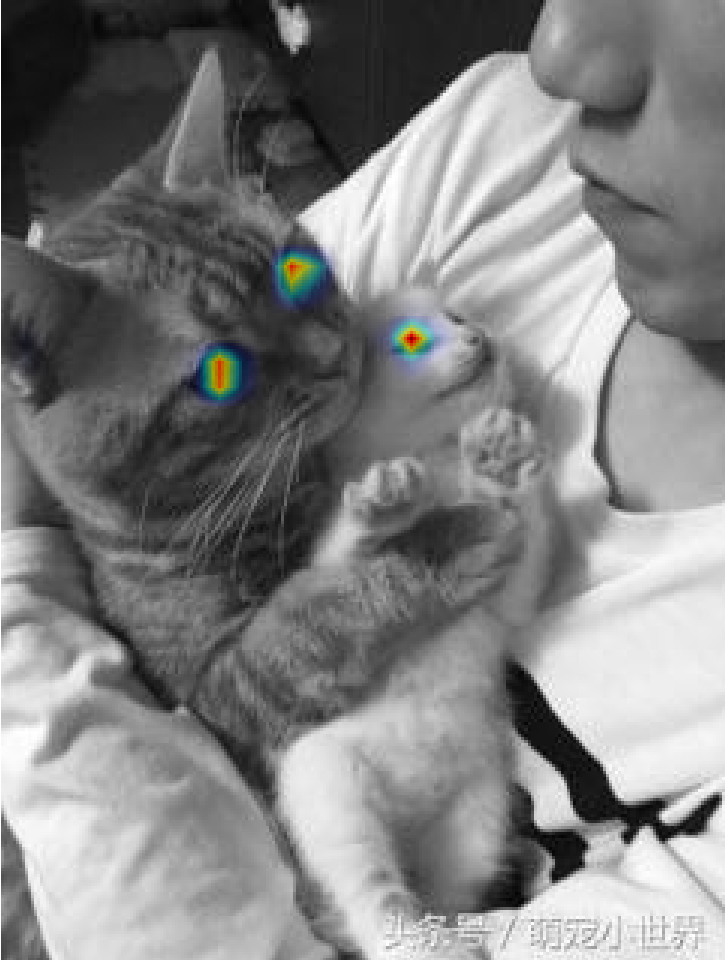} &
		\includegraphics[width=0.18\linewidth]{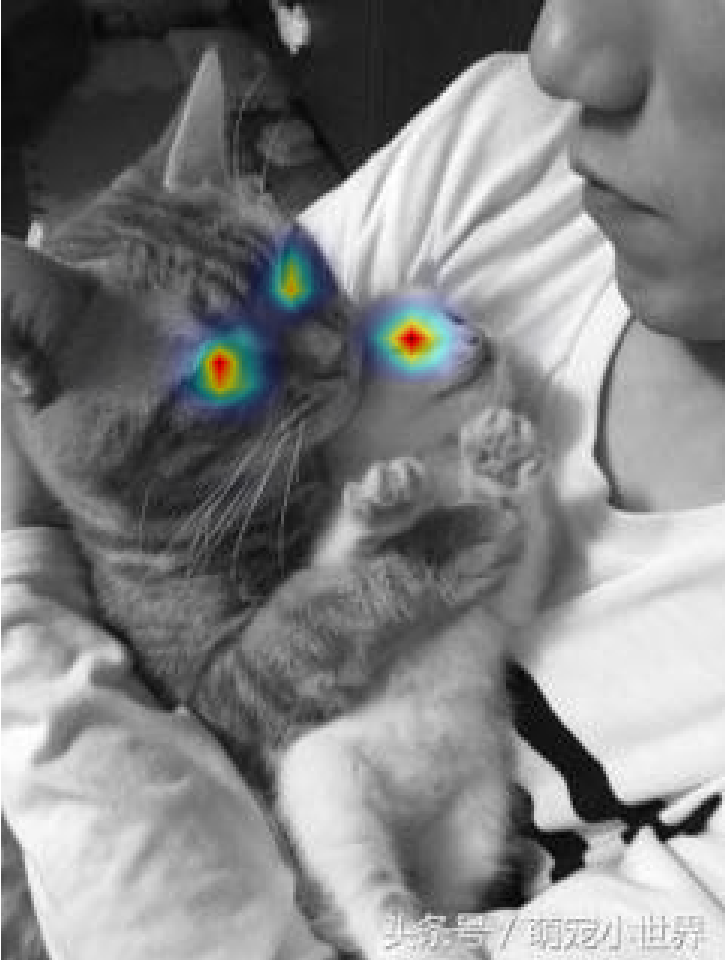} \\ 
	\end{tabular}
	\caption{\small Two sample images to show Sym-NETs performance on out-of-plane reflection and rotation symmetry predictions. The detected reflection symmetry on human nose (bottom row) is a positive surprise, while the miss on the small white cat's face 
shows the limitation of Sym-NETs on detecting small, subtle symmetries.
\label{fig:kittens}} \vspace{-12.5pt}
\end{figure}

\begin{figure*}[!t]
	\centering
	\newlength{\imageRowWidth}
	\setlength{\imageRowWidth}{.087\linewidth}
	\newcommand{\actRow}[3]{
		\includegraphics[width=#3\imageRowWidth]{./results_images/both/extra_images/#1} &
		\includegraphics[width=#3\imageRowWidth]{./results_images/#2/Activations/#1_conv1_0_jet} &
		\includegraphics[width=#3\imageRowWidth]{./results_images/#2/Activations/#1_conv2_0_jet} &
		\includegraphics[width=#3\imageRowWidth]{./results_images/#2/Activations/#1_conv3_0_jet} &
		\includegraphics[width=#3\imageRowWidth]{./results_images/#2/Activations/#1_conv4_0_jet} &
		\includegraphics[width=#3\imageRowWidth]{./results_images/#2/Activations/#1_conv5_0_jet} &
		\includegraphics[width=#3\imageRowWidth]{./results_images/#2/Activations/#1_fc6_sum_jet} &
		\includegraphics[width=#3\imageRowWidth]{./results_images/#2/Activations/#1_fc7_sum_jet}  &
		\includegraphics[width=#3\imageRowWidth]{./results_images/#2/Activations/#1_fc8a_sum_jet}  &
		\includegraphics[width=#3\imageRowWidth]{./results_images/#2/Activations/#1_#2_jet} 
		\\
	}
	\renewcommand{\arraystretch}{.5}
	\setlength{\tabcolsep}{2pt}
	\begin{tabular}{*{10}{c}}
		\multicolumn{10}{c}{\textbf{Reflection Symmetry}} \\
		\parbox{\imageRowWidth}{\centering Original Image} &
		\parbox{\imageRowWidth}{\centering Conv1} &
		\parbox{\imageRowWidth}{\centering Conv2} &
		\parbox{\imageRowWidth}{\centering Conv3} &
		\parbox{\imageRowWidth}{\centering Conv4} &
		\parbox{\imageRowWidth}{\centering Conv5} &
		\parbox{\imageRowWidth}{\centering fc6} &
		\parbox{\imageRowWidth}{\centering fc7} &
		\parbox{\imageRowWidth}{\centering fc8} &
		\parbox{\imageRowWidth}{\centering Sym-NET} \\
		\actRow{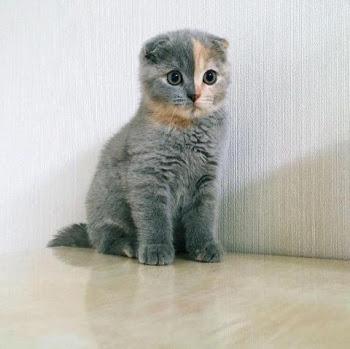}{Reflection}{1}
		\actRow{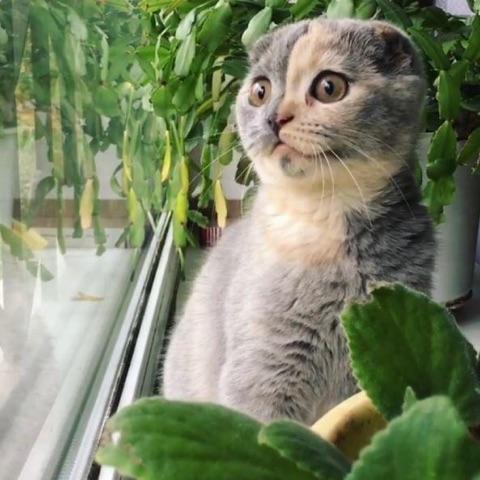}{Reflection}{1}
		\actRow{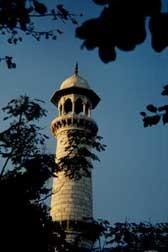}{Reflection}{.7}
		&&&&&&&&&\\
		\multicolumn{10}{c}{\textbf{Rotation Symmetry}} \\
		\parbox{\imageRowWidth}{\centering Original Image} &
		\parbox{\imageRowWidth}{\centering Conv1} &
		\parbox{\imageRowWidth}{\centering Conv2} &
		\parbox{\imageRowWidth}{\centering Conv3} &
		\parbox{\imageRowWidth}{\centering Conv4} &
		\parbox{\imageRowWidth}{\centering Conv5} &
		\parbox{\imageRowWidth}{\centering fc6} &
		\parbox{\imageRowWidth}{\centering fc7} &
		\parbox{\imageRowWidth}{\centering fc8} &
		\parbox{\imageRowWidth}{\centering Sym-NET} \\
		\actRow{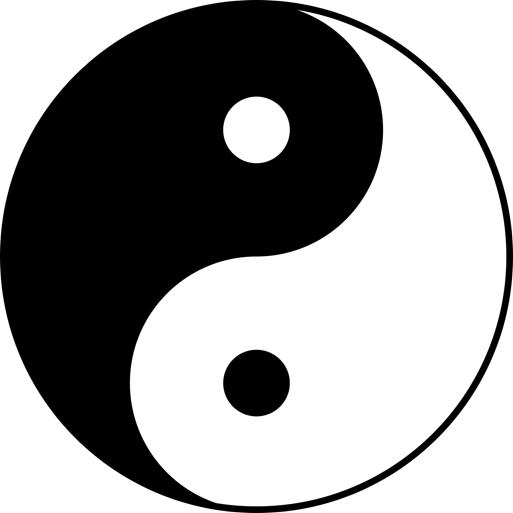}{Rotation}{1}
		\actRow{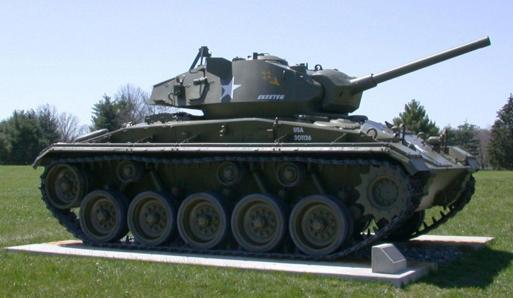}{Rotation}{1}
		\actRow{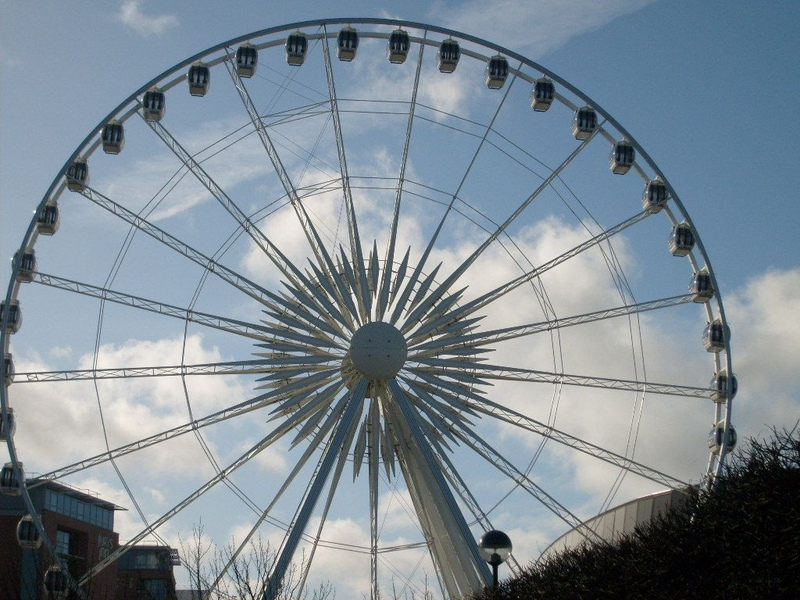}{Rotation}{1}
	\end{tabular}
	\setlength{\tabcolsep}{6pt}
	\caption{Example activations from Sym-VGG showing visualization of the networks. The activations shown are the sum of all the activation channels at each layer. 
	}
	\label{fig:Activations} \vspace{-10pt}
\end{figure*}

\section{Summary and Discussion}

We have shown that our Sym-NETs, trained on human labeled data, can detect mathematically well-defined planar symmetries in images (Figures~\hyperref[fig:PR_ref]{\ref{fig:PR_ref}C}, \hyperref[fig:PR_ref]{\ref{fig:PR_ref}D} CVPR 2013 symmetry detection competition testset), furthermore, it can also capture  
a mixture of {\em symmetries in the wild} that are beyond planar symmetries (Figures~\ref{fig:Montage of Images ref}, \ref{fig:Montage of Images rot}, \ref{fig:kittens}). 
The performance of the Sym-NETs is  significantly superior to existing computer vision algorithms on the test images evaluated 
(Figure \ref{fig:PR_ref}).  
Foreground as well as background symmetries are captured 
(Figures \hyperref[fig:Montage of Images ref]{\ref{fig:Montage of Images ref}B}, \hyperref[fig:Montage of Images ref]{\ref{fig:Montage of Images ref}C}, \hyperref[fig:Montage of Images rot]{\ref{fig:Montage of Images rot}F}, \hyperref[fig:Montage of Images rot]{\ref{fig:Montage of Images rot}G}).

%
Our work has provided an affirmative response to the debate on whether human perception of  
{\em symmetry in the wild}  can be  computationally modeled, and the deep-learning platform offers us a means to do so. 
However, this is only an encouraging beginning.  The questions of WHAT features are learned, and HOW multiple-visual, spatial  and/or semantic cues are combined to achieve the superior performance of Sym-NET remain. 
By peeking into the inner layers of activations in the Sym-NETs (Figure \ref{fig:Activations}), we observe that for reflection symmetry, the color/shading cues fade away at deeper layers in promoting the reflection axis; for rotation symmetry, local cues seem to contribute much more to rotation centers than global (or distant) ones. 
%
%
Some observed human-machine discrepancies that lower the Sym-NET performance include: 
\begin{tightitemize}
\item Size: Humans are better in detecting small (rotation) symmetries, e.g. the clock in Figure \hyperref[fig:symmetry basics]{\ref{fig:symmetry basics} BOTTOM}), while Sym-NET fails  (Figures~\hyperref[fig:Montage of Images rot]{\ref{fig:Montage of Images rot}C}, \hyperref[fig:Montage of Images rot]{\ref{fig:Montage of Images rot}E})
	\item Subtlety: Humans are keen at perceiving object-symmetry that is barely visible from the background, e.g. a laptop computer, Figure \hyperref[fig:Montage of Images ref]{\ref{fig:Montage of Images ref}E}, Sym-NETs can miss such subtleties Figure~\hyperref[fig:kittens]{\ref{fig:kittens} BOTTOM}
		\item Humans do not consistently label the same semantic object (such as eyes) while the networks learn to predict eyes as rotationally symmetric reliably: e.g. Figure \hyperref[fig:Montage of Images rot]{\ref{fig:Montage of Images rot}C} and Figure \hyperref[fig:symmetry basics]{\ref{fig:symmetry basics} TOP} (dog eyes).
	
\end{tightitemize}


It has been widely accepted that symmetry perception serves as a mid-level cue that is 
important to human understanding of the world, ranging from how to combine shapes together into objects~\cite{parovel2002mirror}, to identify foreground from background~\cite{driver1992preserved}, and to judge attractiveness~\cite{scheib1999facial}.  
Therefore, computer vision problems such as semantic segmentation, image understanding, scene parsing, and 3D reconstruction may benefit greatly from reliable characterizations of symmetry in the data. 
After many years of practice, it is about time we question the robustness of those computer vision algorithms that are solely based on first principles (i.e. mathematical definition of symmetry), and open up to a {\em hybridisation} of modern computing technology with classic theory. Our initial experiment with Sym-NETs has set an optimistic  starting point.  
More examples and resources can be found on our project website: \url{http://vision.cse.psu.edu/research/beyondPlanarSymmetry/index.shtml}.


\section{Acknowledgement}
 
This work is supported in part by an NSF CREATIV grant (IIS-1248076). 
\vfill
\pagebreak
{\small
	\bibliographystyle{ieee}
	\bibliography{paper,combined_citation_final_sorted}
}

\vfill

\end{document}